\documentclass[10pt,twocolumn,letterpaper]{article}

\usepackage{iccv}
\usepackage{times}
\usepackage{epsfig}
\usepackage{graphicx}
\usepackage{amsmath}
\usepackage{amssymb}
\usepackage{epstopdf}
\usepackage[ruled,vlined]{algorithm2e}

\iccvfinalcopy % *** Uncomment this line for the final submission

\def\iccvPaperID{1540} % *** Enter the ICCV Paper ID here

\ificcvfinal\fi

\begin{document}

%%%%%%%%% TITLE
\title{Convolutional Neural Pyramid for Image Processing}

\author{Xiaoyong Shen \quad Ying-Cong Chen \quad Xin Tao \quad Jiaya Jia\\
The Chinese University of Hong Kong\\
{\tt\small \{xyshen, ycchen, xtao, leojia\}@cse.cuhk.edu.hk}}

\maketitle
%\thispagestyle{empty}

%%%%%%%%% ABSTRACT
\begin{abstract}

We propose a principled convolutional neural pyramid (CNP) framework for general
low-level vision and image processing tasks. It is based on the essential finding that
many applications require large receptive fields for structure understanding. But
corresponding neural networks for regression either stack many layers or apply large
kernels to achieve it, which is computationally very costly. Our pyramid structure can
greatly enlarge the field while not sacrificing computation efficiency. Extra benefit
includes adaptive network depth and progressive upsampling for quasi-realtime testing on
VGA-size input. Our method profits a broad set of applications, such as depth/RGB image
restoration, completion, noise/artifact removal, edge refinement, image filtering, image
enhancement and colorization.
\end{abstract}

%%%%%%%%% BODY TEXT
\section{Introduction}

Convolutional neural networks (CNNs) become the most powerful tool for high-level
computer vision tasks, such as object detection \cite{Girshick15}, image classification
\cite{KrizhevskySH12} and semantic segmentation \cite{LongSD15}. Along this line, CNNs
were also used to solve image processing problems. Representative applications are
image/video super-resolution
\cite{DongLHT16_SRCNN,DongLT16,KimLL15a,KimLL15b,LiaoTLMJ15}, artifact and noise removal
\cite{DongDLT15,EigenKF13,MaoSY16a}, completion/inpainting \cite{RenXYS15,XieXC12},
learning image filtering \cite{XuRYLJ15,LiuP016}, image deconvolution \cite{XuRLJ14},
\etc

Compared to very successful classification and detection, low-level-vision neural
networks encounter quite different difficulties when they are adopted as {\it regression}
tools.

\begin{figure}[t]%%
\centering
\includegraphics[width=0.88\linewidth]{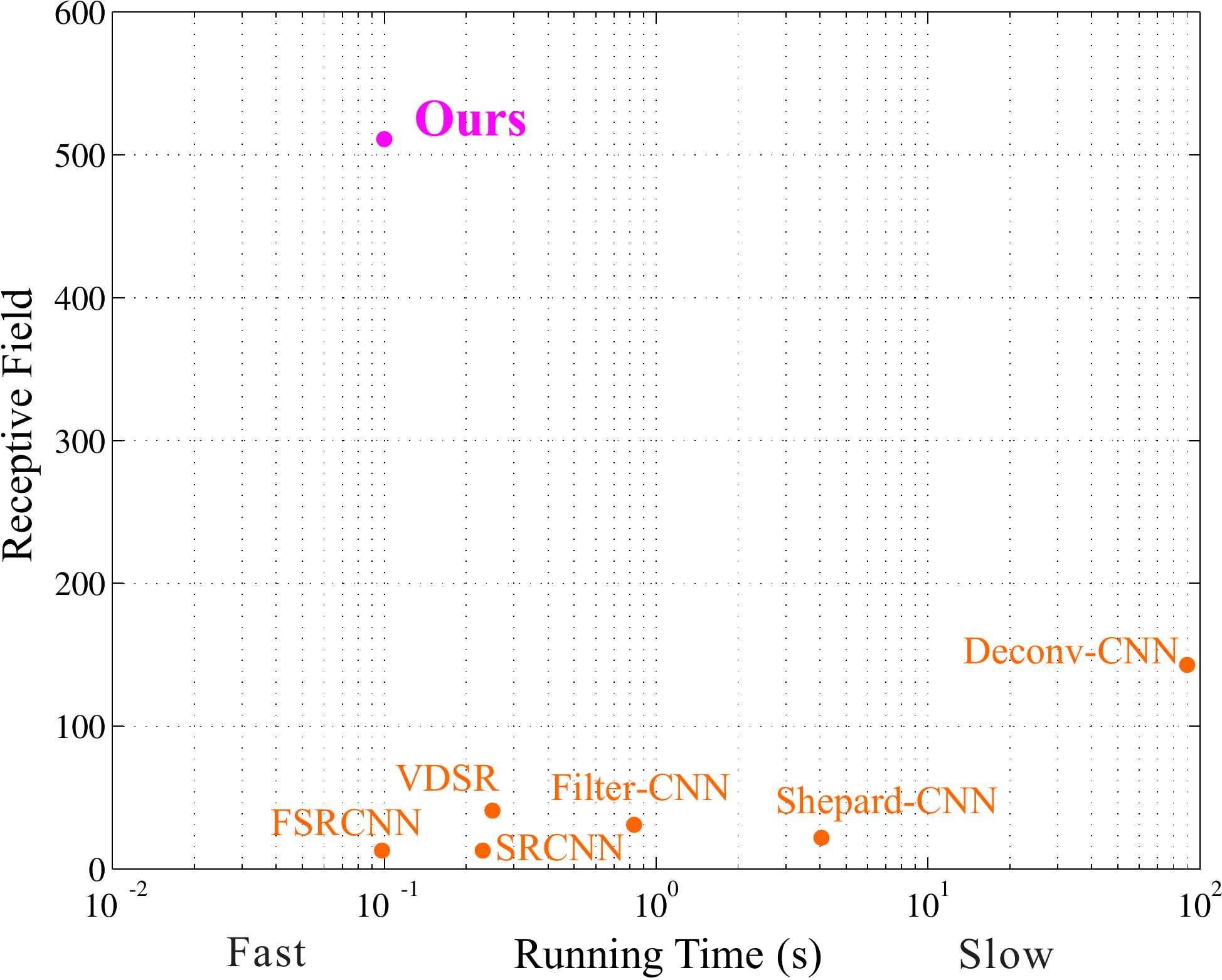}\\
\caption{Receptive field size vs. running time of current low-level-vision CNNs,
including SRCNN \cite{DongLHT16_SRCNN}, Filter-CNN \cite{XuRYLJ15}, VDSR \cite{KimLL15b},
Shepard-CNN \cite{RenXYS15} and Deconv-CNN \cite{XuRLJ14}. Our network achieves a very
large receptive field efficiently, which enables many general learning applications.
Running time is obtained during testing on a Nvidia Titan X graphics card with VGA-size
color-image input.\vspace{-0.1in}} \label{fig:diff_receptive_field}
\end{figure}

\begin{figure*}[t]%%
\centering
\begin{tabular}{@{\hspace{0.0mm}}c@{\hspace{1.0mm}}c@{\hspace{1.0mm}}c@{\hspace{1.0mm}}c@{\hspace{0mm}}}
\includegraphics[width=0.24\linewidth]{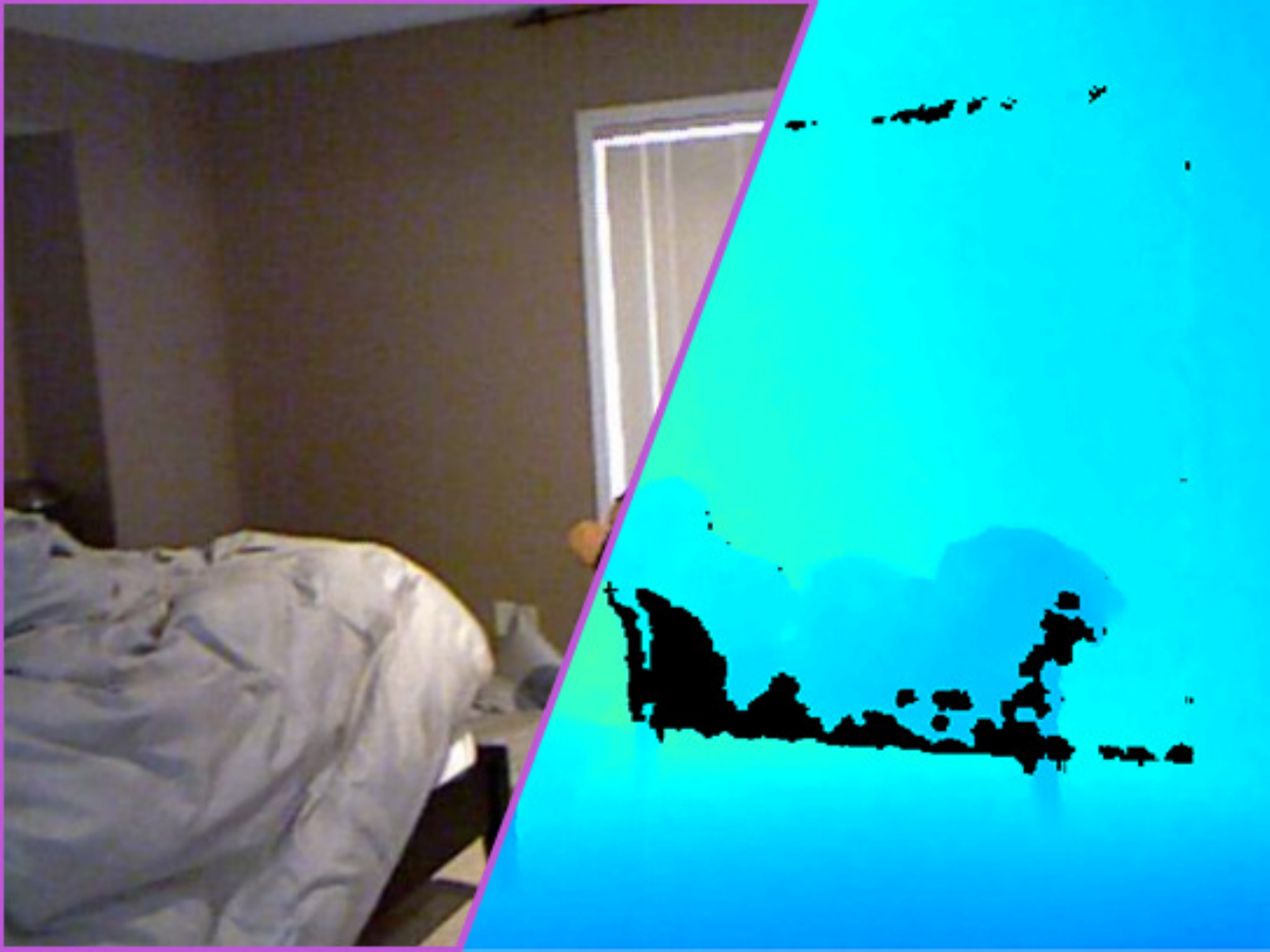}&
\includegraphics[width=0.24\linewidth]{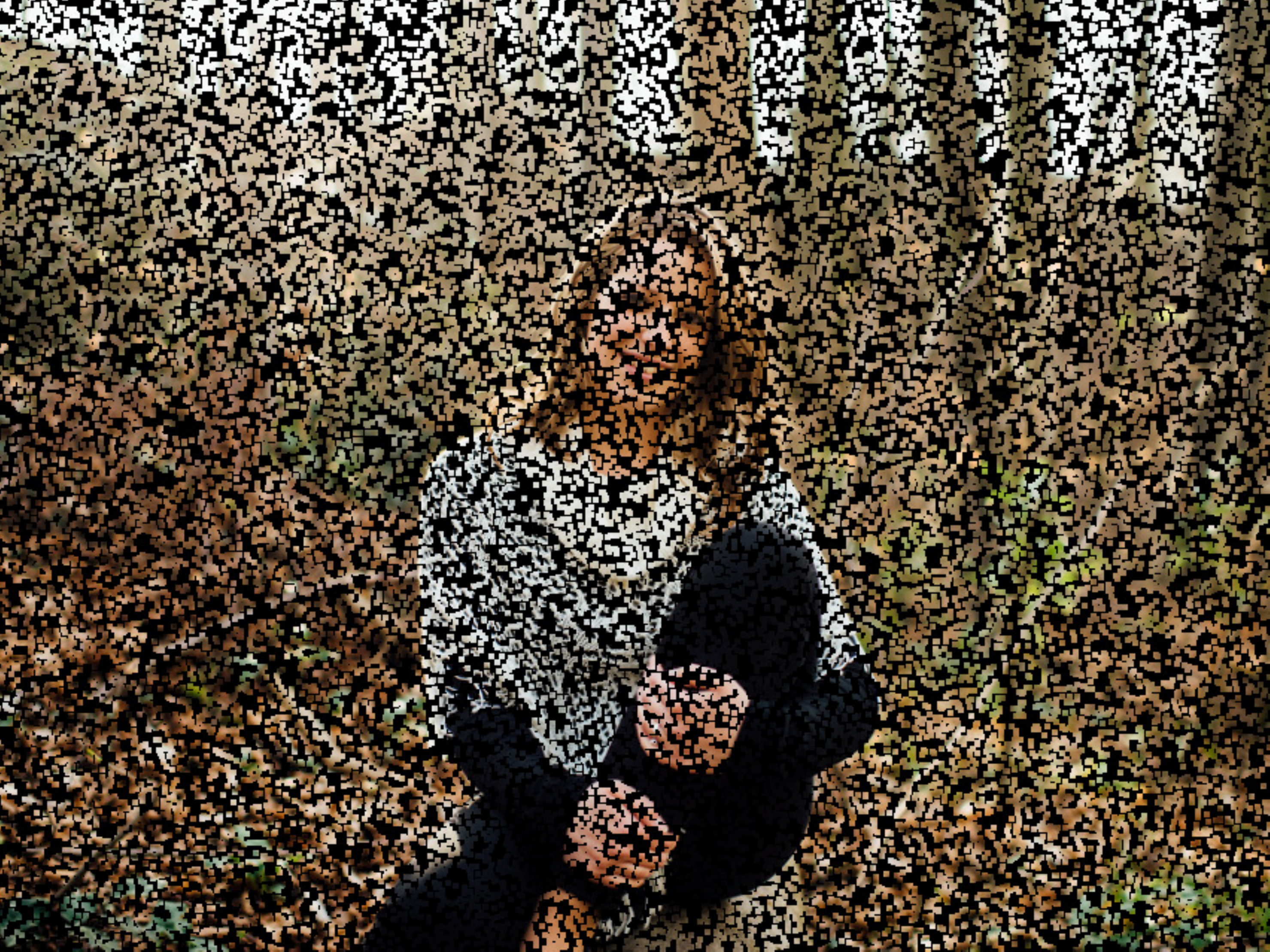}&
\includegraphics[width=0.24\linewidth]{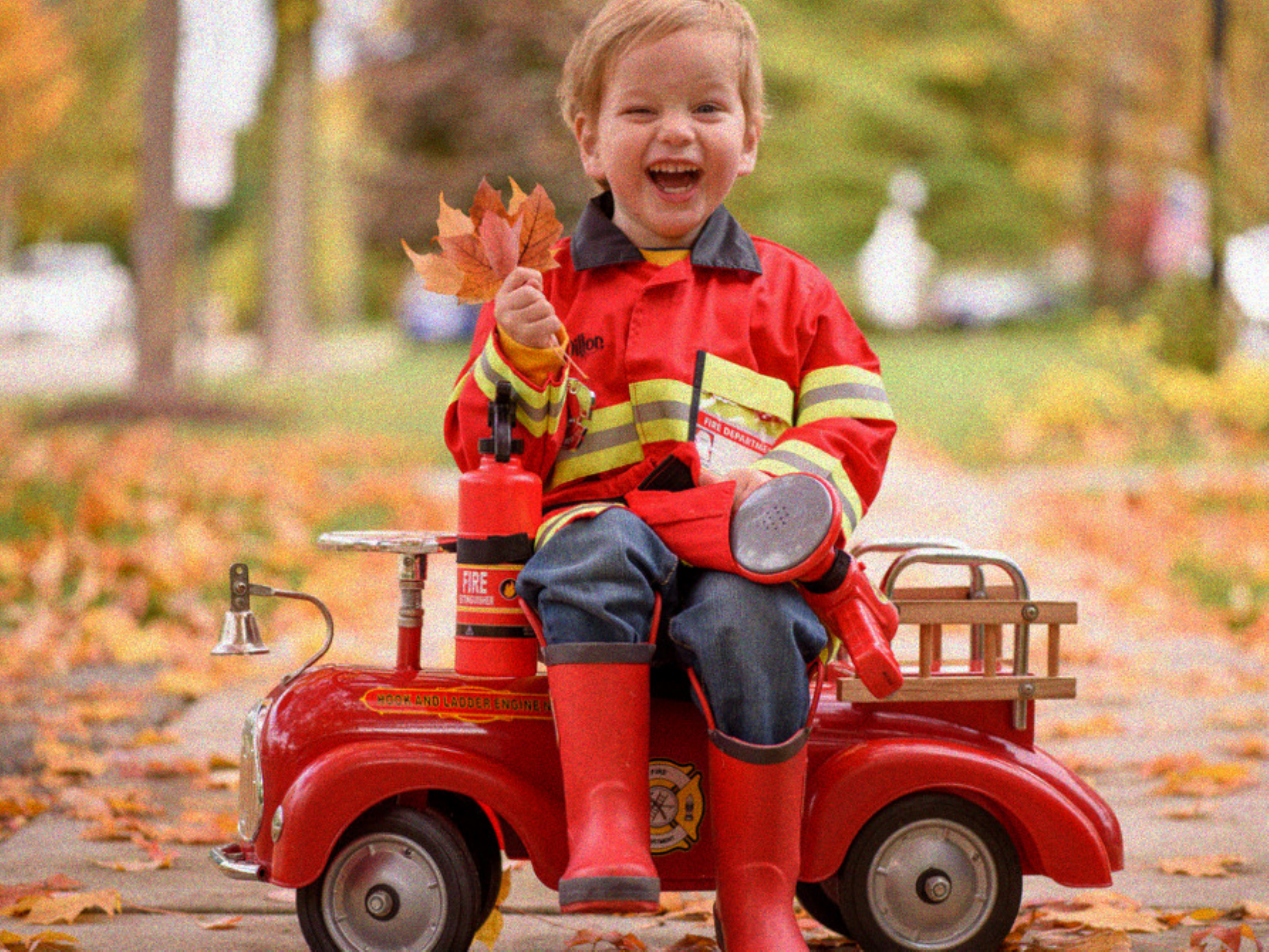}&
\includegraphics[width=0.24\linewidth]{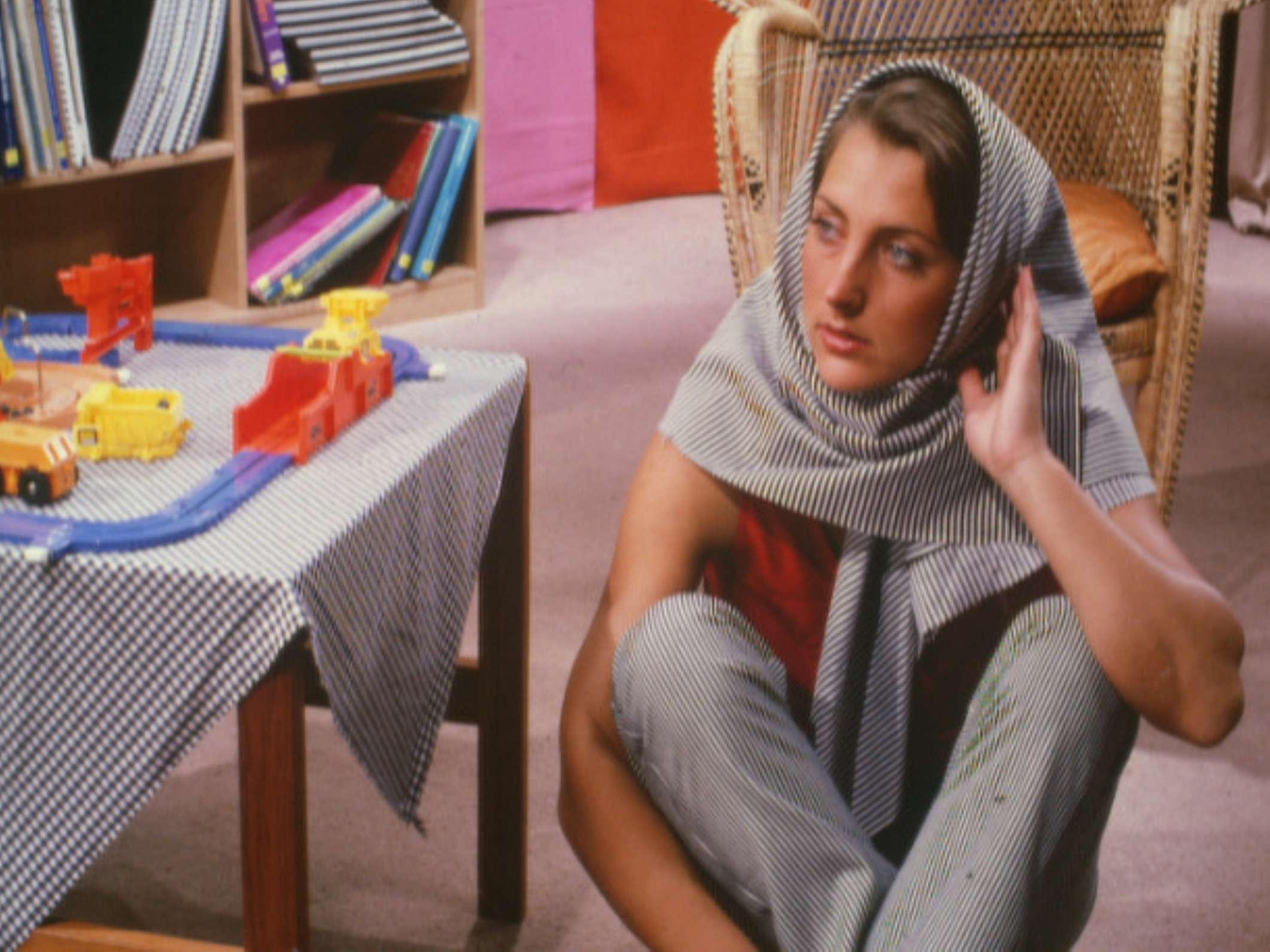}\\
\includegraphics[width=0.24\linewidth]{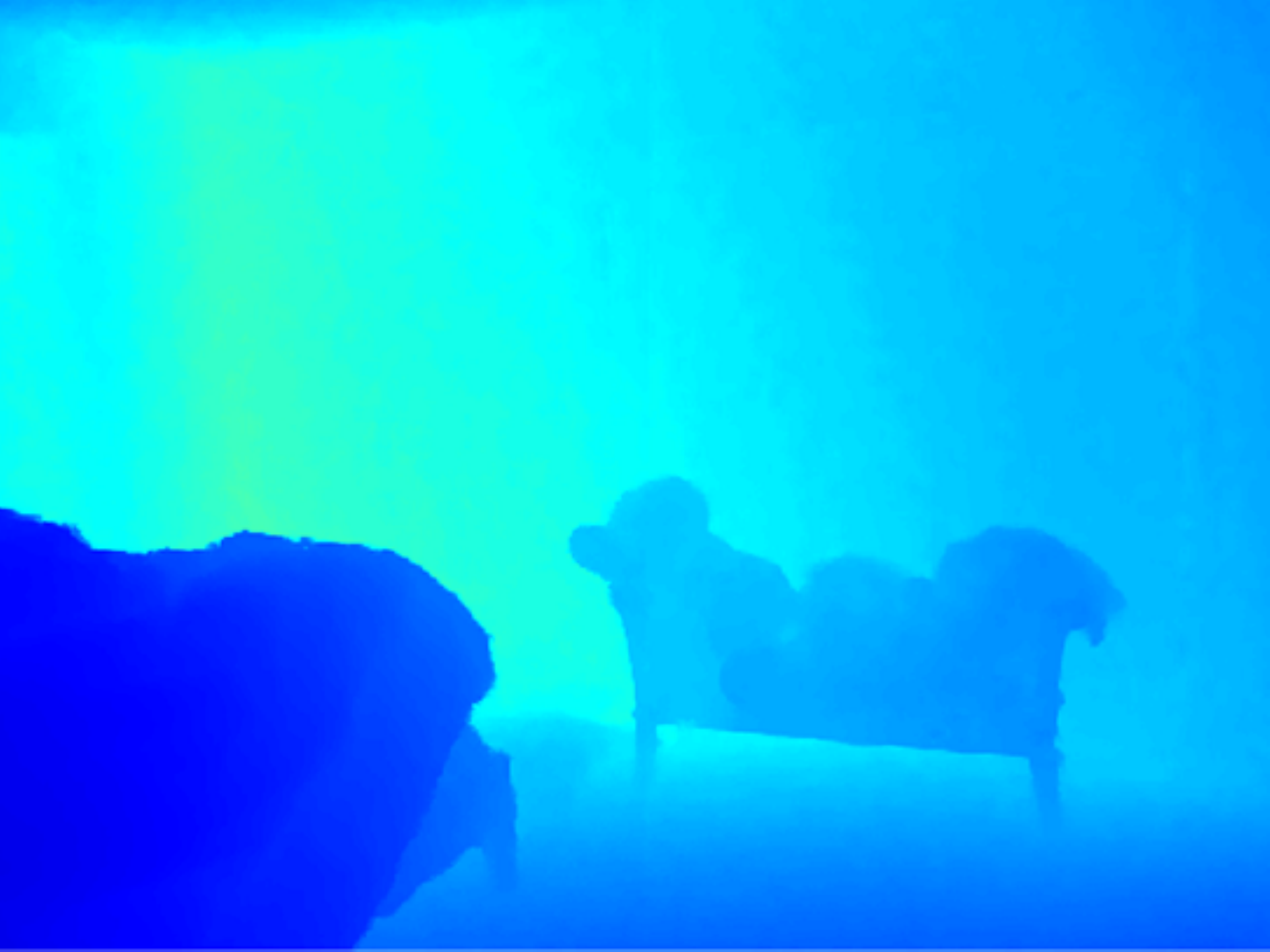}&
\includegraphics[width=0.24\linewidth]{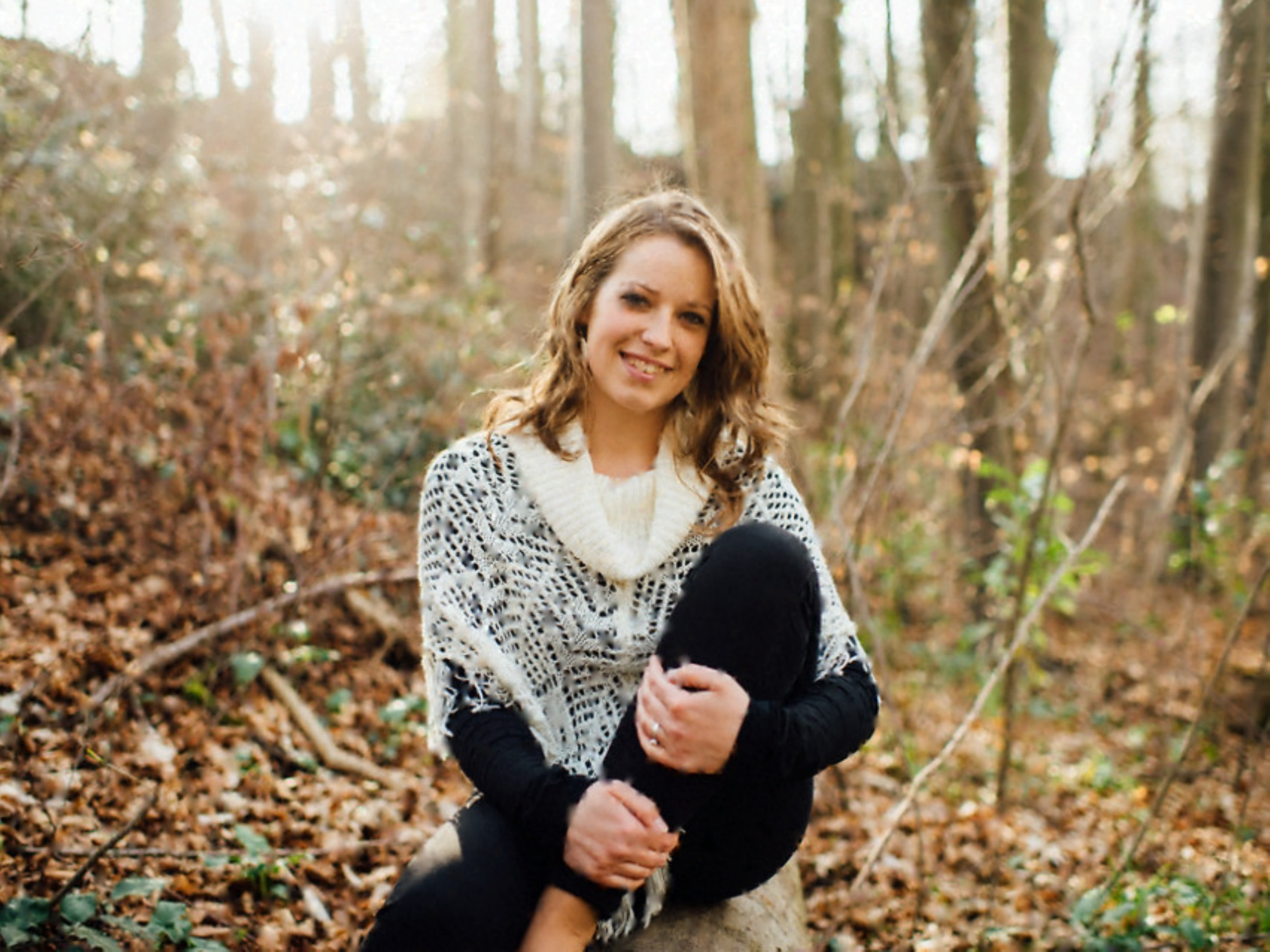}&
\includegraphics[width=0.24\linewidth]{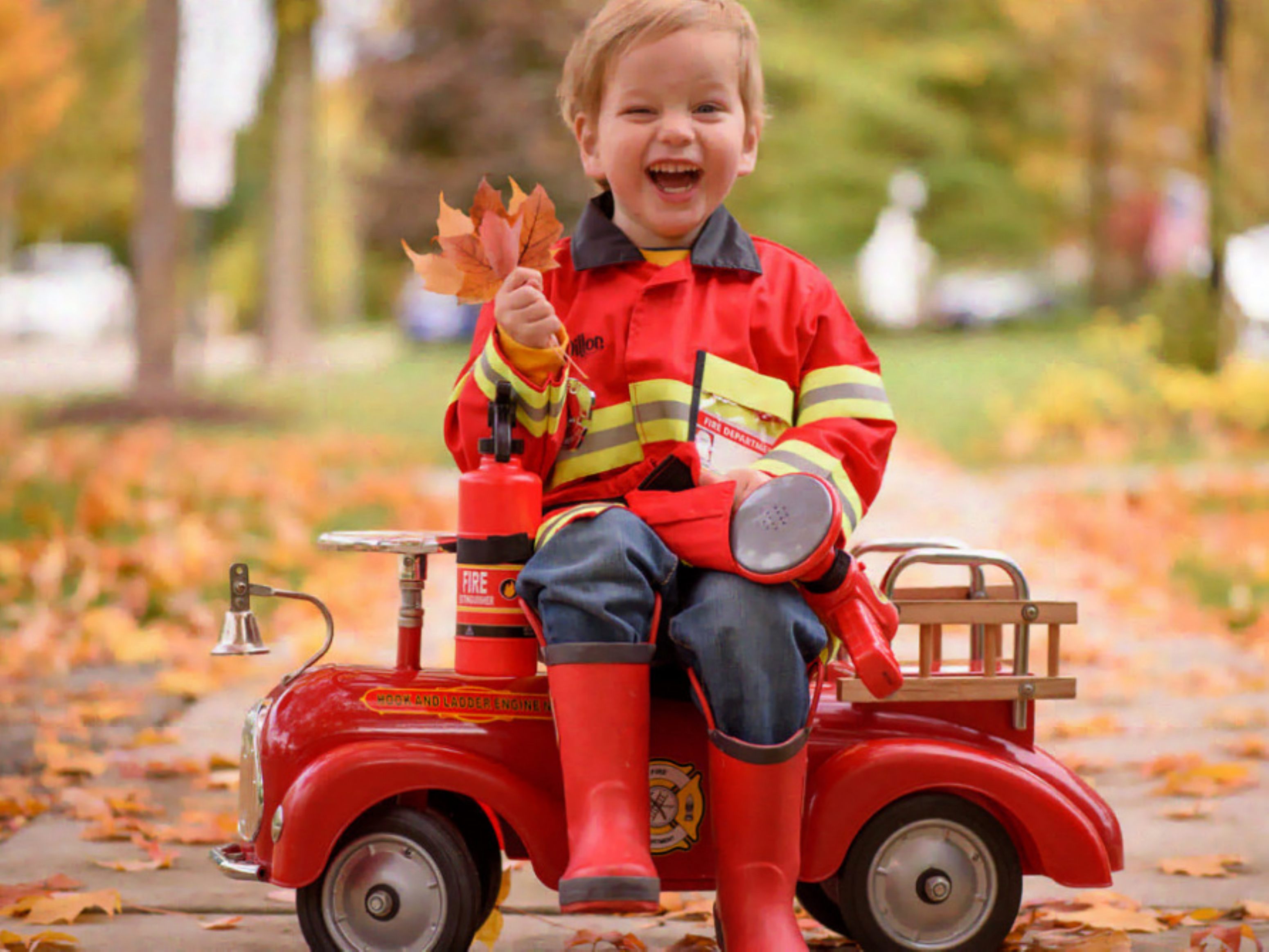}&
\includegraphics[width=0.24\linewidth]{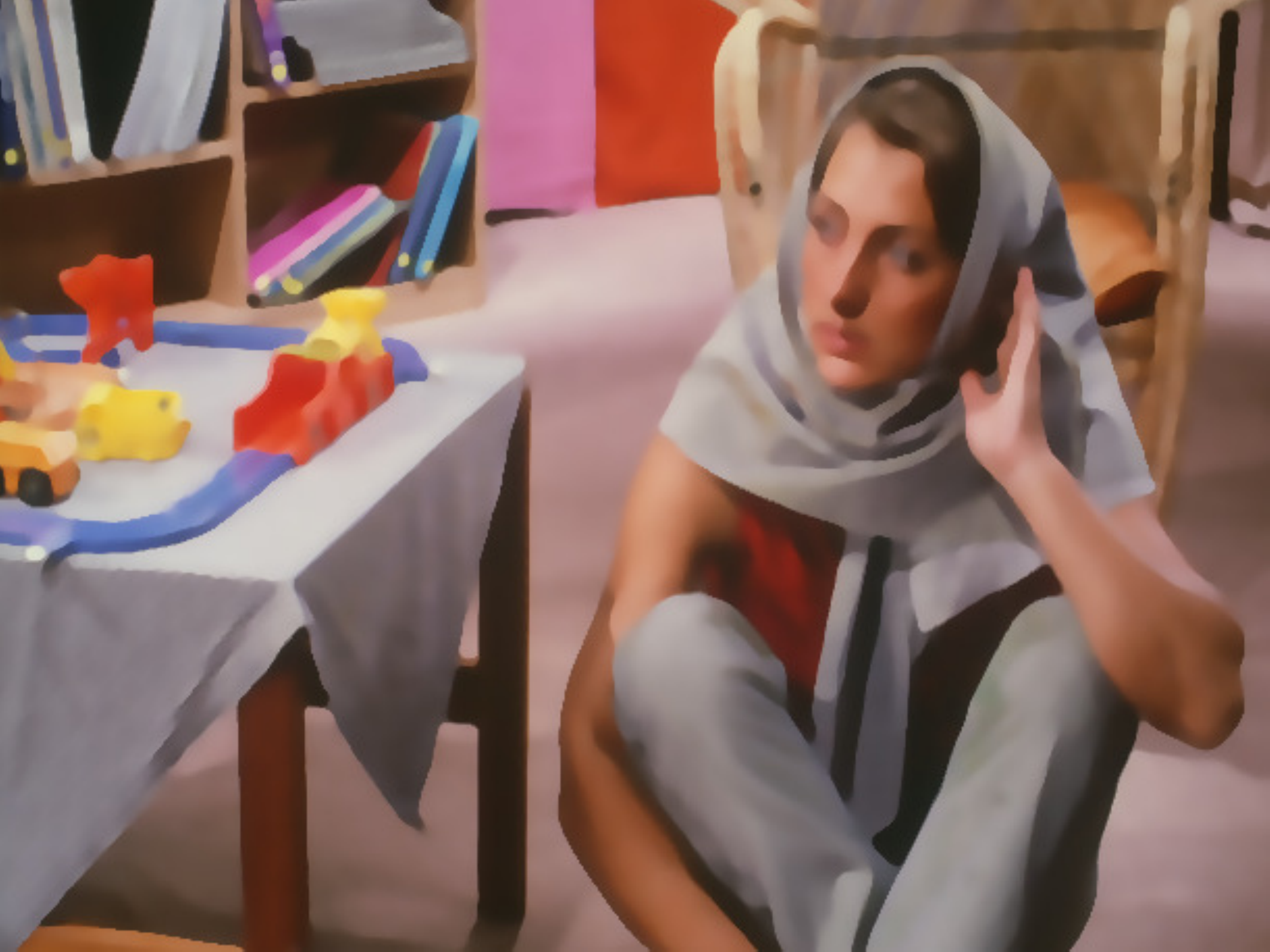}\\
\small (a) Depth/RGB Restoration & \small (b) Image Completion & \small (c) Artifact/Noise Removal & \small (d) Learning Image Filter \\
\end{tabular}
\caption{Our convolutional neural pyramid benefits many image-level applications
including depth/RGB image restoration, image completion, artifact/noise removal, and
learning image filter.\vspace{-0.1in}} \label{fig:application}
\end{figure*}

\vspace{-0.1in}\paragraph{Size of Receptive Field for Image Processing} The first problem
is about the receptive field, which is the region in the input layer connected to an
output neuron. A reasonably large receptive field can capture global information applied
to inference. In VGG \cite{SimonyanZ14a} and ResNet \cite{HeZRS15}, a large receptive
field is achieved mainly by stacking convolution and pooling layers. For low-level-vision
CNNs \cite{DongLHT16_SRCNN,XuRLJ14,KimLL15b,XuRYLJ15,LiHA016,DongLT16}, the pooling
layers are commonly removed in order to regress the same-resolution output and preserve
image details. To still obtain acceptable receptive fields with the convolution layers,
the two solutions are to use large kernels (\eg $7\times7$ or $9\times9$)
\cite{DongLHT16_SRCNN,XuRYLJ15,LiHA016} and stack many layers \cite{KimLL15b}.

We found unexceptionally these two schemes both make the system run slowly due to heavy
computation and consume a lot of memory. To illustrate it, we show the relation between
the size of a receptive field and running time of current low-level-vision CNNs in Figure
\ref{fig:diff_receptive_field}. It is observed based on this plot that most existing
low-level-vision CNNs, such as SRCNN \cite{DongLHT16_SRCNN}, Filter-CNN \cite{XuRYLJ15},
VDSR \cite{KimLL15b}, and Shepard-CNN \cite{RenXYS15}, achieve receptive fields smaller
than $41\times41$ (pixels). Their applications are accordingly limited to
local-information regression ones such as super-resolution, local filtering, and
inpainting. On the other hand, Deconv-CNN \cite{XuRLJ14} uses $143\times143$ field size.
But its computation cost is very high as shown in Figure \ref{fig:diff_receptive_field}.

Contrary to these common small receptive fields, our important finding is that {\it large
or even whole-image fields are essential for many low-level vision tasks} because most of
them, including image completion, restoration, colorization, matting, global filtering,
are based on global-information optimization.

In this paper, we address this critical and general issue, and design a new network
structure to achieve very-large receptive fields without sacrificing much computation
efficiency, as illustrated in Figure \ref{fig:diff_receptive_field}.

\begin{figure}[t]
\centering
\includegraphics[width=0.96\linewidth]{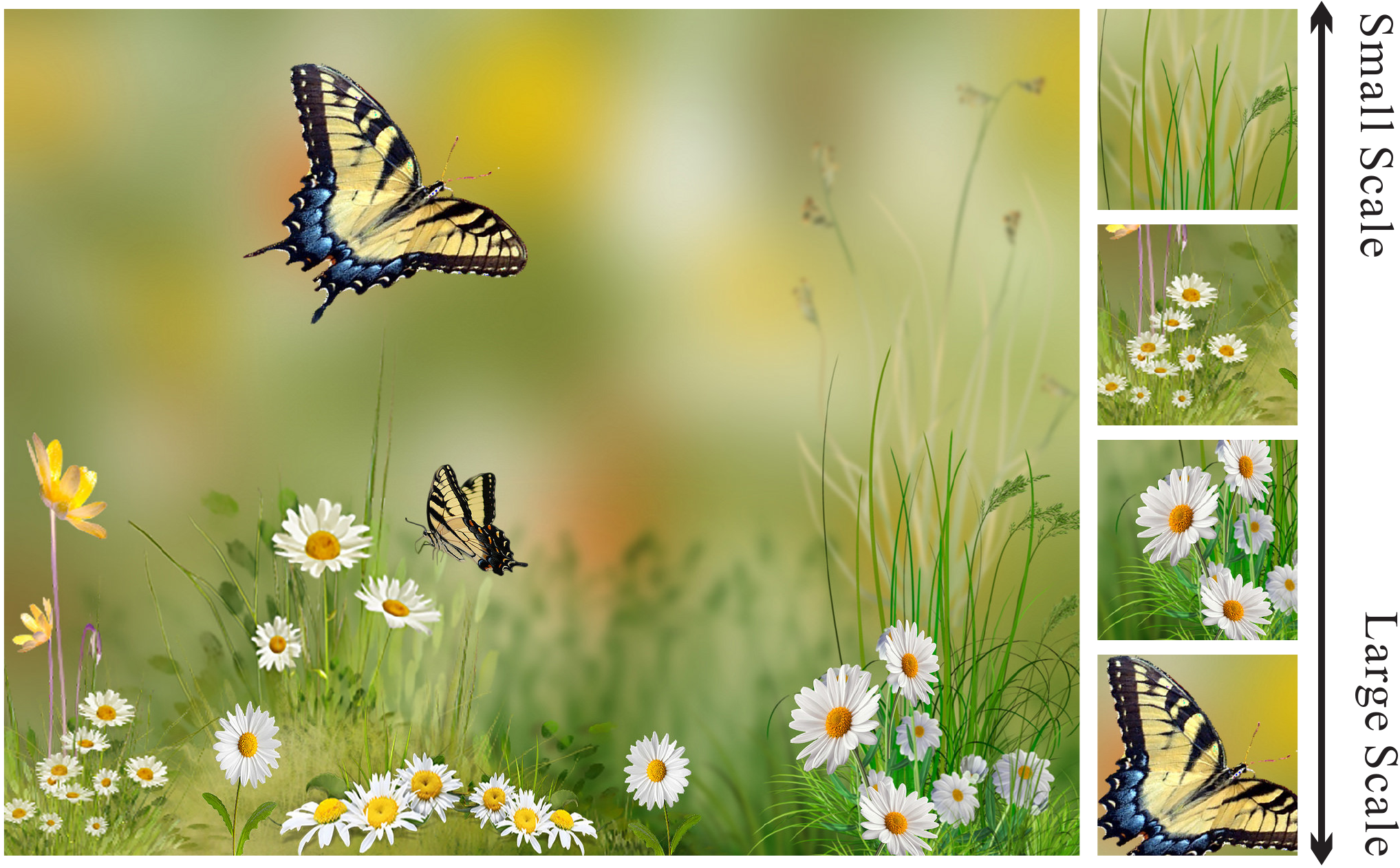}
\caption{Natural images contain different scales of details, texture, and objects.
\vspace{-0.1in}} \label{fig:diff_scale}
\end{figure}

\vspace{-0.1in}\paragraph{Information Fusion} The second difficulty is on multiscale
information fusion. It is well known that most image content is with multiscale patterns
as shown in Figure \ref{fig:diff_scale}. As discussed in \cite{ZeilerF14}, the
small-scale information such as edge, texture and corners are learned in early layers and
object-level knowledge comes from late ones in a deep neural network. This analysis also
reveals the fact that {\it color and edge information vanishes in late hidden layers}.
For pixel labeling work, such as semantic segmentation \cite{LongSD15} and edge detection
\cite{LiuP016}, early feature maps were extracted and taken into late layers to improve
pixel inference accuracy.

We address this issue differently for low-level-vision tasks that do not involve pooling.
Although it seems that the goal of retaining early-stage low-level edge information
contradicts large receptive field generation in a deep network, our solution shows they
can accomplished simultaneously with a general network structure.

\begin{figure*}[t]
\centering
\includegraphics[width=0.99\linewidth]{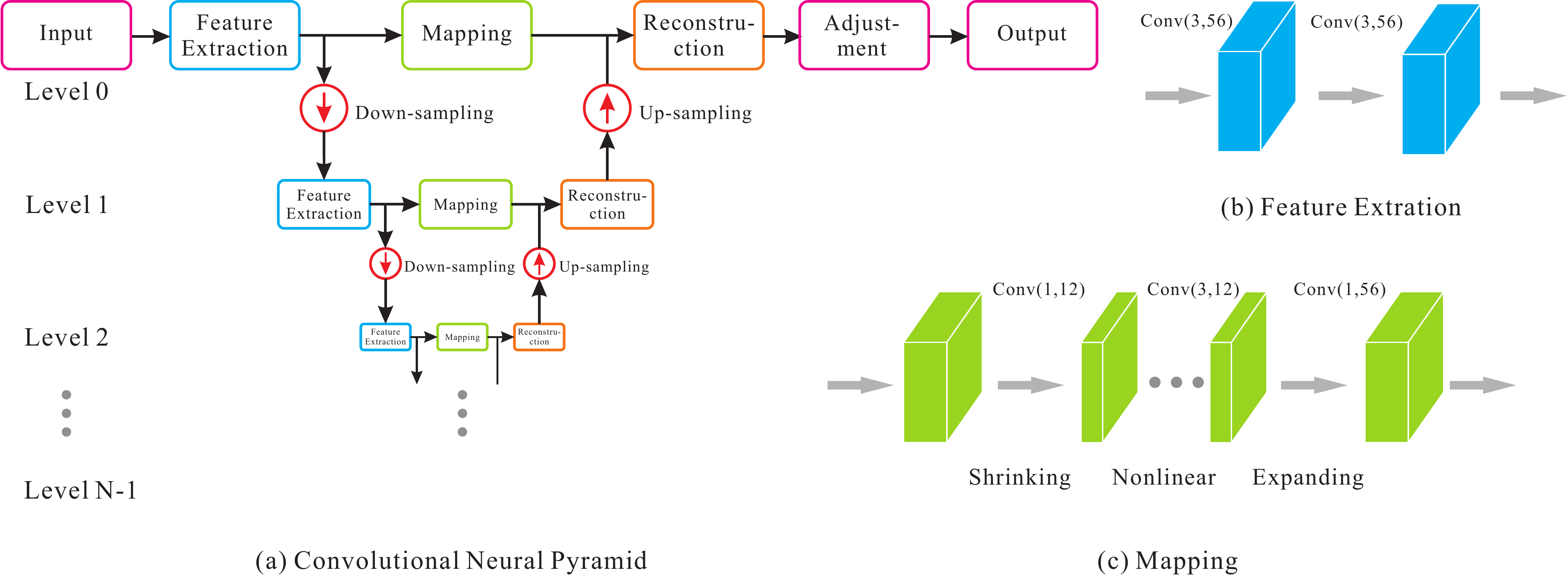}
\caption{Illustration of our convolutional neural pyramid. (a) shows the convolutional
pyramid structure. (b) and (c) are the feature extraction and mapping components
respectively. $Conv(x,y)$ denotes the convolution operation, where $x$ is the kernel size
and $y$ is the number of output.\vspace{-0.1in}} \label{fig:illustration}
\end{figure*}

\vspace{-0.1in}\paragraph{Our Approach and Contribution} To address aforementioned
difficulties, we propose {\it convolutional neural pyramid (CNP)}, which can achieve
quite large receptive fields while not losing efficiency as shown in Figure
\ref{fig:diff_receptive_field}. It intriguingly enables multiscale information fusion for
general image tasks. The convolutional neural pyramid structure is illustrated in Figure
\ref{fig:illustration} where a cascade of features are learned in two streams. The first
stream across different pyramid levels plays an important role for enlarging the
receptive field. The second stream learns information in each pyramid level and finally
merges it to produce the final result.

Our CNP benefits a wide spectrum of applications. As demonstrated in Figure
\ref{fig:application}, we achieve the state-of-the-art performance in depth/RGB restoration,
image completion/inpainting, noise/artifact removal, \etc Other applications include
learning image filter, image colorization, optical flow and stereo map refinement, image
enhancement, and edge refinement. They are shown in the experiment section and our
supplementary file.

Our framework is very efficient in computation. On an Nvidia Titan X display card, 28
frames are processed per second for a QVGA-size input and 9 frames per second for a VGA-size input for {\it all} image tasks.

\section{Related Work}

We review recent progress of using CNNs for computational photography and low-level
computer vision.

\vspace{-0.1in}\paragraph{Inverse Problems} CNNs were used to solve inverse problems such
as image super-resolution/upsampling and deconvolution. Representative methods include
SRCNN \cite{DongLHT16_SRCNN}, very-deep CNNs \cite{KimLL15b}, deeply-recursive CNNs
\cite{KimLL15a}, and FSRCNN \cite{DongLT16}. The major scheme is to regress the inverse
degradation function by stacking many CNN layers, such as convolution and ReLU. They are
mostly end-to-end frameworks. Image super-resolution frameworks were extended to
video/multi-frame super-resolution \cite{LiaoTLMJ15} and depth upsampling \cite{HuiLT16}.
All these frameworks are with relatively small receptive fields, which are hard to be
extended to general image processing tasks.

Another important topic is image deconvolution, which needs image priors, such as
gradient distribution, to constrain the solution. Xu \etal \cite{XuRLJ14} demonstrated
that simple CNNs can well approximate the inverse kernel and achieved decent results. As
shown in Figure \ref{fig:diff_receptive_field}, large kernels for convolution require
heavy computation.

\vspace{-0.15in}\paragraph{Decomposition} Image decomposition applies filtering. Xu \etal
learned general image filters of bilateral \cite{TomasiM98}, rolling guidance
\cite{ZhangSXJ14}, RTV \cite{XuYXJ12}, $L_0$ \cite{XuLXJ11}, etc. For learned recursive
filter Liu \etal \cite{LiuP016}, image structure/edge information is inferred and fed
into the recursive layers as weights. This framework is efficient. Deep joint image
filtering \cite{LiHA016} addresses the joint estimation problem.

Besides filtering, CNNs were also used for compression artifact removal \cite{DongDLT15},
dirt/rain removal \cite{EigenKF13}, denoising \cite{XieXC12,MaoSY16a}, \etc Simply put,
these frameworks employ CNNs to regress image structure. We show that our deep
convolutional pyramid can better address global-optimization-like decomposition problems
because of our large receptive fields with information fusion.

\vspace{-0.15in}\paragraph{Image Completion/Inpainting} Considering sparse coding and
CNNs, an auto-encoder framework for blind image inpainting was proposed in
\cite{XieXC12}. K\"{o}hler \etal \cite{KohlerSSH14} directly trained a CNN taking a
specified mask and origin image as input. Ren \etal \cite{RenXYS15} enhanced the
convolution layer to Shepard convolution. These methods also produce limited receptive
fields. Recently, a semantic inpainting framework \cite{YehCLHD16} incorporates
perceptual and contextual loss. They achieved good results even with large holes for
similar-content images.

\vspace{-0.15in}\paragraph{Representative CNNs for High-level Tasks} Many CNNs are
proposed for high-level recognition tasks. Famous ones include AlexNet
\cite{KrizhevskySH12}, VGG \cite{SimonyanZ14a}, GoogleLeNet \cite{SzegedyLJSRAEVR15} and
ResNet \cite{HeZRS16}. Based on these frameworks, lots of networks are proposed to solve
semantic image segmentation as pixel classification. They include FCN \cite{LongSD15},
DeepLab \cite{ChenPK0Y16}, SegNet \cite{BadrinarayananH15}, U-Net \cite{RonnebergerFB15}
and those of \cite{YuK15,LinMS016,GhiasiF16,PohlenHML16}. Compared with these
image-to-label or image-to-label-map frameworks, our CNP aims for image processing
directly by image-to-image regression.

\section{Convolutional Neural Pyramid}

To achieve large receptive fields, our convolutional neural pyramid network is
illustrated in Figure \ref{fig:illustration}. It includes levels from $0$ to $N-1$ as
shown in Figure \ref{fig:illustration}(a) where $N$ is the number of levels. We denote
these levels as $L_i$ where $i\in \{0\cdot\cdot\cdot N-1\}$. Different-scale content is
encoded in them. The feature extraction, mapping and reconstruction operations are
denoted as $\mathcal{F}_i$, $\mathcal{M}_i$ and $\mathcal{R}_i$ respectively in each
pyramid level. The input to $L_i$ is the feature extracted from $L_{i-1}$ after
downsampling, essential to enlarge the receptive-field.

Specifically, the input to $L_0$ is the raw data, such as RGB channels. After feature
extraction and mapping in each level, information is reconstructed from $L_{N-1}$ to
$L_0$ where the output of each $L_i$ is upsampled and fused with the output from
$L_{i-1}$. Thus scale-sensitive information is used together to reconstruct the final
output.

\subsection{Beyond Simple Multiscale Fusion}
\label{sec:simplefus} Before going to details for each component, we explain that our CNP
is fundamentally different from and superior to the intuitive multiscale fusion strategy
regarding the overall structure where the latter scheme processes each-scale input by
feature extraction, mapping and reconstruction respectively and sums these outputs into
the final result. There are two major differences in our network structure.

\vspace{-0.15in}\paragraph{Adaptive Network Depth in Levels} First, in the original
resolution, i.e., level 0 in Figure \ref{fig:illustration}(a), only two convolution
layers are performed for feature extraction. It is a shallow network that can well
capture the edge and color information as previously discussed -- its complexity is low.

Then in level 1, after downsampling to reduce feature map size, another feature extraction
module is performed, which involves more convolution layers for deeper processing of the
input image. It begins to extract more complicated patterns beyond edges due to its
higher depth and accordingly larger receptive fields. The complexity is not much
increased with the reduced feature map size.

When the network contains more levels, they are with even smaller feature maps processed more
deeply to form semantically meaningful information regarding objects and shapes. The
receptive fields are further increased while the computation complexity is caped to a
reasonably small value. This {\it adaptive depth} strategy is remarkably effective in a
lot of tasks. We will show experimental evaluation later.

\vspace{-0.15in}\paragraph{Progressive Upsampling}

During the final upsampling phase (after mapping) in Figure \ref{fig:illustration}(a),
directly upsampling the output from level $i$ to level $0$ needs a large kernel since the
upsampling ratio is $2^i$, which makes learning difficult and possibly inaccurate. We
address this issue by a progressive scheme from level $N-1$ to $0$. For each phase, the
output in level $i$ is upsampled and then fused with the output in level $i-1$. Thus, the
ratio is only 2 where $3\times3$ efficient upsampling kernel suffices. Further,
information in level $i-1$ is upsampled with the guidance from level $i$ since the
upsampling kernel is learned between the two neighboring levels. It avoids possible edge
and feature distortion when reconstructing the final result.

The empirical comparison with the simple multiscale fusion strategy will be provided
after explaining all following components in our network.

\subsection{Component Details}

\paragraph{Feature Extraction} We apply convolution layers to extract
image features. As shown in Figure \ref{fig:illustration}(b), two convolution layers with
PReLU rectification are employed for each feature extraction module. The low-level image
features such as edges and corners can be extracted by one module \cite{ZeilerF14} in
level 0. Semantically more meaningful information is constructed with more extraction
modules regarding our adaptive depth in other levels.

Each convolution layer outputs $56$ features with a $3\times3$ kernel. For level 0, its
feature output is a 56-channel map. Other levels take the input of feature extracted from
another level after downsampling. Thus, $2*(i+1)$ convolution layers are applied to
feature extraction in level $L_i$, where $i$ is the index, to learn complicated features.
This pyramid design effectively enlarges the receptive field without quickly increasing
computation. It also nicely encodes different-scale image information.

\vspace{-0.15in}\paragraph{Mapping} The extracted features in each level $L_i$ are
transformed by mapping $\mathcal{M}_i$. As demonstrated in Figure
\ref{fig:illustration}(c), our mapping includes shrinking, nonlinear transform and
expanding blocks, motivated by the work of \cite{DongLT16}. The shrinking layer embeds
the 56-channel feature map into 12 channels by $1\times 1$ convolution. It reduces the
feature dimension and benefits noise reduction.

The network is then stacked by nonlinear transform layers. Instead of applying
large-kernel convolution \cite{XuRLJ14,DongLHT16_SRCNN,LiHA016} to achieve large
receptive fields, we stack $S$ convolution layers with kernel size $3\times3$ to reduce
parameters and empirically achieve comparable performance. Since nonlinear transform is
conducted in a low-dimension feature-space, computation efficiency is further improved.
Here $S$ ranges from $1-3$. We will discuss its influence later.

Finally, an expanding layer is added to remap the 12-channel feature map into 56 channels
by $1\times 1$ convolution, for reverse of the shrinking operation. We use this layer to
gather more information to reconstruct high quality results.

\vspace{-0.15in}\paragraph{Reconstruction} The reconstruction operation fuses information
from two neighboring levels. For $L_i$ and $L_{i+1}$, the output of $L_{i+1}$ is
upsampled and then fused with the output from $L_{i}$. The goal is to merge
different-scale information progressively. We test two fusion operations of concatenating
two outputs and element-wise sum of the outputs. They achieve comparable performance in
our experiments. We thus use element-wise sum for a bit higher efficiency.

\vspace{-0.15in}\paragraph{Down- and Up-sampling} Down- and up-sampling are key
components in pyramid construction and collapse. The downsampling operation shown in
Figure \ref{fig:illustration}(a) resizes extracted feature maps. For simplicity, we only
consider resizing ratio $0.5$ in our network. Two downsampling schemes are tested -- max
pooling and convolution with a $3\times 3$ kernel with stride $2$. The simple max pooling
works better in experiments due to preservation of the max response. Note that in level
$0$, our network does not include any pooling layer to keep image details. We simply
implement the upsampling operation as a deconvolution layer in Caffe \cite{JiaSDKLGGD14}.

\vspace{-0.15in}\paragraph{Adjustment} Since the reconstructed feature maps in $L_0$ are
with 56 channels, to adjust difference between the feature map and output, we use two
convolution layers with kernel size $3\times 3$ to generate the same number of channels
as the final output, as shown in Figure \ref{fig:illustration}(a). Before each
convolution layer, a PReLU layer is applied.

Algorithm \ref{alg:deeppyramid} gives the overall procedure, $\mathcal{F}_i(\mathbf{X})$
and $\mathcal{M}_i(\mathbf{X})$ are output of $\mathbf{X}$ after operations
$\mathcal{F}_i$ and $\mathcal{M}_i$ respectively.
$\mathcal{R}_i(\mathbf{X},\mathbf{X}_1)$ is the reconstruction result of $\mathbf{X}$ and
$\mathbf{X}_1$ with operation $\mathcal{R}_i$. Configuration of our network is provided
in the supplementary file.

\begin{algorithm}[t]
%    \SetAlgoLined \LinesNumbered  \DontPrintSemicolon
    \SetKwInOut{Input}{INPUT}\SetKwInOut{Output}{OUTPUT}
    \Input{$\mathbf{X}$}
    \Output{$\mathbf{Y}$}
    \BlankLine
    /*Feature Extraction and Mapping*/ \;
    $\mathbf{F}_0 \gets \mathcal{F}_0(\mathbf{X})$ \;
    \For {t:= 1 {\bf to} $N$ }{
        $\mathbf{F}_{t} \gets \mathcal{M}_t(\mathcal{F}_t(\downarrow \mathbf{F}_{t-1})) $ \quad /*$\downarrow$ is down-sampling*/
    }
    $\mathbf{F}_0 \gets \mathcal{M}_0(\mathbf{F}_0)$ \;
    \BlankLine
    /*Reconstruction*/ \;
    \For {t:= $N-1$ {\bf to} $0$ }{
        $\mathbf{F}_{t} \gets \mathcal{R}_t( \mathbf{F}_{t}, \uparrow \mathbf{F}_{t+1}) $
        \quad /*$\uparrow$ is up-sampling*/
    }
    \BlankLine
    $\mathbf{Y} \gets \mathbf{F}_0$ \;
    \caption{Convolutional Neural Pyramid}
    \label{alg:deeppyramid}
\end{algorithm}

\subsection{More Analysis}\label{sec:analysis}

The quite large receptive field, adaptive depth, and progressive fusion achieved by our
CNP greatly improve performance for many tasks. We perform extensive evaluation and give
the following analysis.

\vspace{-0.1in}\paragraph{Receptive Field vs. Running Time~~} In our $N$-level network,
the information passing through level $L_{N-1}$ has the largest receptive field. For this
level, the images are processed by feature extraction and downsampling for $N$ passes,
which introduce a total of $2*N$ feature extraction operations. The receptive field size
increases exponentially with the number of levels, making the largest receptive field
$2^N$ times of the original one. Note that increasing the pyramid level introduces
limited extra computation since the size of feature map decreases quickly in levels. The
total extra computation for level ${N-1}$ is only $1/2^N$ of that in level $0$ for
feature extraction.

The receptive field and running time statistics are reported in Table
\ref{tab:timeSpace}. For 5 levels, our testing time is only about 3 times of that spent
in level 0 including overhead in up- and down-sampling. Note that there is only one
feature extraction module in level 0 while level 4 has 5 of them. Our effective receptive
field is as large as $511\times511$ pixels.

We compare our CPN model with a baseline ``Single-Level CNN" in Table \ref{tab:timeSpace}
that does not contain any pyramid levels -- so its structure is similar to level 0 of our
model. In order to get $95\times95$ receptive field, 48 $3 \times 3$ convolution layers
are needed, which consumes nearly 9GB GPU memory. Similar performance is yielded for our
3-level system.

\begin{table}[t]
\centering \setlength{\tabcolsep}{.07cm} \footnotesize
\begin{tabular}{|l|c|c|c|c|c|c|}
  \hline
  & \multicolumn{3}{c|}{Our CPN} & \multicolumn{3}{c|}{Single-Level CNN} \\
  \hline
  Recept. Field & Levels & Time (s) & Mem. (GB) & Layers & Time (s) & Mem. (GB) \\
  \hline
  \hline
   $15 \times 15$ & 1 & 0.034 & 2.66 & 8 & 0.034 & 2.66\\
  \hline
  $39 \times 39$ & 2 & 0.089 & 3.48 & 20 & 0.102 & 4.58\\
  \hline
  $95 \times 95$ & 3 & 0.104 & 3.70 & 48 & 0.252 & 8.96\\
  \hline
  $223 \times 223$ & 4 & 0.109 & 3.76 & 112 & N/A & $>12$\\
  \hline
  $511 \times 511$ & 5 & 0.111 & 3.77 & 256 & N/A &  $>12$\\
  \hline
\end{tabular}\vspace{0.1in}
\caption{Time and memory consumption analysis in terms of different sizes of receptive
fields. ``Single-Level CNN" denotes the baseline model without pyramid levels. VGA-size
input images are used in experiments. Our GPU memory does not allow 112 or 256 layers.
}\label{tab:timeSpace}
\end{table}

\begin{table}[t]
\centering \small
\begin{tabular}{|l|c|c|}
  \hline
  Total Levels~~~~& ~Simple multiscale Fusion~ & ~~~~~Ours~~~~~ \\
  \hline
  \hline
  $2$ & 34.41 & 34.66\\
  \hline
  $3$ & 36.83 & 36.94\\
  \hline
  $4$ & 37.09 & 37.99\\
  \hline
  $5$ & 38.41 & 39.36\\
  \hline
\end{tabular}\vspace{0.1in}
\caption{Comparison with simple multiscale fusion on NYU Depth V2 data
\cite{SilbermanHKF12}. PSNRs are reported for each setting. }\label{tab:multiscale}
\end{table}

\begin{table}[t]
\centering \small
\begin{tabular}{|c|c|c|c|}
  \hline
  & ~~~$\mathcal{S}=1$~~~ & ~~~$\mathcal{S}=2$~~~ & ~~~$\mathcal{S}=3$~~~ \\
  \hline
  \hline
  $N=0$ & 32.07 & 32.31 & 32.67\\
  \hline
  $N=1$ & 34.31 & 34.66 & 35.01\\
  \hline
  $N=2$ & 36.71 & 36.94 & 37.09\\
  \hline
  $N=3$ & 37.75 & 37.99 & 38.01\\
  \hline
  $N=4$ & 38.98 & 39.36 & 39.42\\
  \hline
\end{tabular}\vspace{0.1in}
\caption{Effectiveness of our method. PSNRs are reported according to the numbers of
pyramid levels ($N$) and nonlinear transform layers ($S$) on the dataset of
\cite{SilbermanHKF12}.}\label{tab:parachange}
\end{table}

\begin{figure*}[t]
\centering
\includegraphics[width=0.93\linewidth]{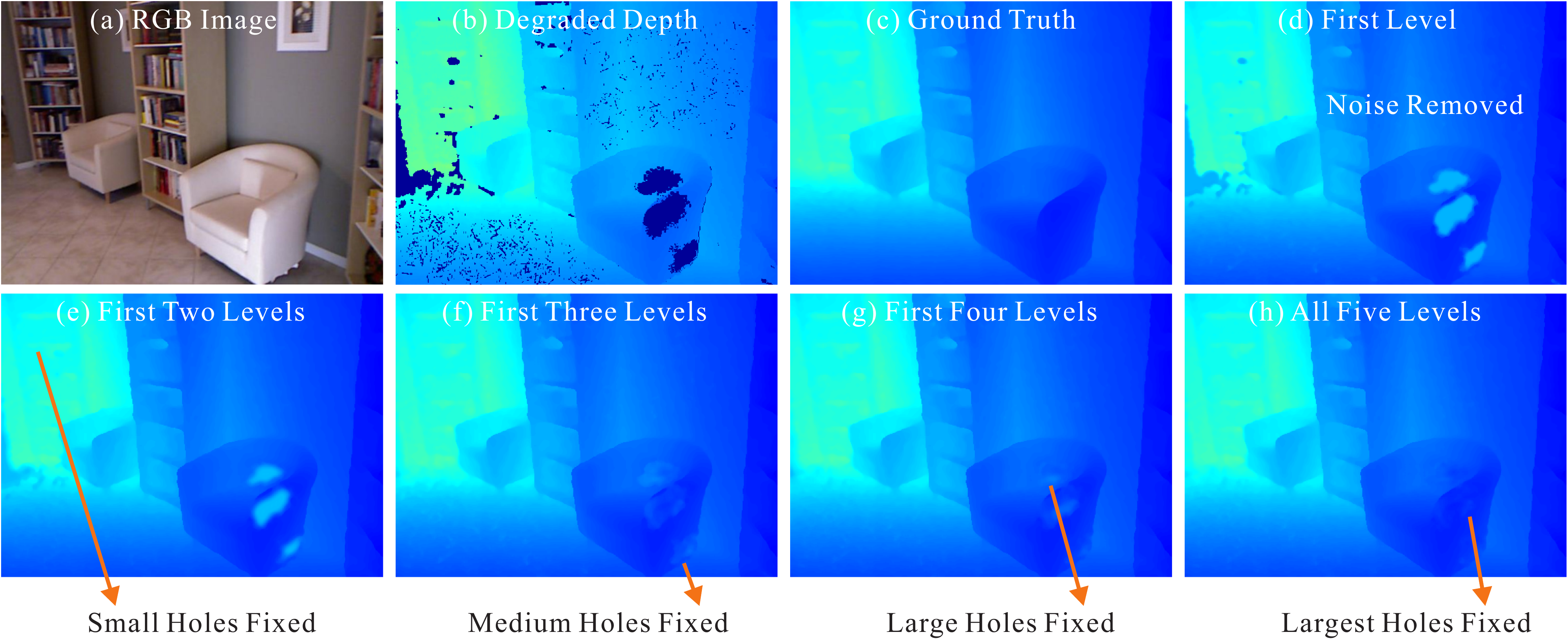}
\caption{Effectiveness of our convolution neural pyramid. (a) and (b) are the reference
RGB image and corresponding degraded depth map. (c) is the ground truth depth. (d-h) show
results of different pyramid levels.\vspace{-0.1in}} \label{fig:effectiveness}
\end{figure*}

\vspace{-0.1in}\paragraph{Comparison with Simple Multiscale Fusion} We compare our
framework with simple multiscale fusion described in Section \ref{sec:simplefus} under
the same parameter setting and report the results in Table \ref{tab:multiscale} for
depth/RGB image restoration on NYU Depth V2 benchmark \cite{SilbermanHKF12}. Our
framework outperforms this alternative under all levels. More pyramid levels yield larger
improvement because we adapt network depth in levels while simple multiscale fusion does
not.

\vspace{-0.1in}\paragraph{Effectiveness of the Pyramid} We use the same depth/RGB
restoration benchmark to verify the effectiveness of our network because this task needs
large-region information in different scales to fill in holes and remove artifacts. More
details of the task are included in Section \ref{sec:experiment}. We evaluate our network
under different setting and report the results in Table \ref{tab:parachange}. They
indicate involving several pyramid levels can greatly improve depth/RGB restoration
quality since it benefits from afore discussed large receptive field sizes. We also test
the influence of $\mathcal{S}$, which is the number of the nonlinear transform layers.
Setting it to $1-3$ does not significantly change the result since the receptive field is
not much influenced.

\vspace{-0.15in}\paragraph{Visualizing What CNP Learns} We visualize the information CNP
learns regarding its different pyramid levels for depth restoration. We block some-level
information to understand how it proceeds. The model is with $5$ levels and $1$ transform
layer each. We first block levels $L_1-L_4$ and only show the result inferred from $L_0$
in Figure \ref{fig:effectiveness}(d) -- the small receptive field of $L_0$ is good to
remove small holes. For large ones, artifacts are generated.

Then we gradually include more levels and show results in Figure
\ref{fig:effectiveness}(e-h). It is obvious that the capacity of removing holes increases
as the total pyramid level expands. The biggest region on the chair is completely
restored when all 5 levels are used. This manifests that these pyramid levels capture
different degrees of semantic information. They work best on particular structure
respectively. Our fusion process makes good use of all of them.

\vspace{-0.15in}\paragraph{Relation to Previous Methods~~} Traditional convolution
pyramids \cite{FarbmanFL11} is only to approximate large-kernel convolution by a series
of small ones. It is limited to gradient interpolation and does not extend to learning
easily. Our network is obviously different due to the new structure and ability to learn
general low-level vision operations.

When the number of our pyramid level is one, our network is similar to the network for
image super-resolution \cite{DongLT16}. We note our contribution is exactly on the
pyramid structure and the effective adaptive depth for large receptive field generation.
It can handle a large variety of complex tasks, which the method of \cite{DongLT16} does
not. We show these applications in what follows.

\vspace{-0.15in}\paragraph{Difference from Detection and Segmentation CNNs~~} Compared
with popular CNNs such as VGG \cite{SimonyanZ14a}, ResNet \cite{HeZRS15} and FCN
\cite{LongSD15}, the key difference is that our framework directly handles image-to-image
{\it regression for image processing} in a very efficient way. Although some semantic
segmentation frameworks, such as RefineNet \cite{LinMS016}, U-Net \cite{RonnebergerFB15},
and FRRN \cite{PohlenHML16}, also consider multi-scale information, they are
fundamentally different due to the diverse nature of classification and regression. None
of these methods can be directly applied (or through slight modification with regression
loss) to our low-level vision tasks.

\section{Experiments and Applications} \label{sec:experiment}

We implement our network in Caffe \cite{JiaSDKLGGD14} - training and testing are on a
single NVIDIA Titan X graphics card. For most applications, we set $N=5$ and
$\mathcal{S}=1$ by default. Following \cite{KimLL15b}, we perform residual learning for
all tasks. During training, $81\times81$ patch size and $32$ batch size are applied. We
use learning rate $10^{-5}$ for all model training. In general, training with 2 to 8
epoches is enough.

We set the training loss as the combination of $L_2$ loss in intensity and gradient. The
reason is that intensity loss retains appearance while the gradient loss has the ability
to keep results sharp. The loss is expressed as $ \mathcal{L}(X,\hat{X}) = \|X-\hat{X}\|
+ \lambda \|\nabla X-\nabla \hat{X}\|$, where $X$ is the prediction and $\hat{X}$ is the
ground truth. $\nabla$ is the gradient operator and $\|\cdot\|$ computes the $L_2$ norm
distance. $\lambda$ is the parameter balancing the two losses, set to $1$ in all
experiments. Our framework can also use other loss functions, such as the generative
loss. Exploring them will be our future work.

\begin{figure*}[t]
\centering
\includegraphics[width=0.98\linewidth]{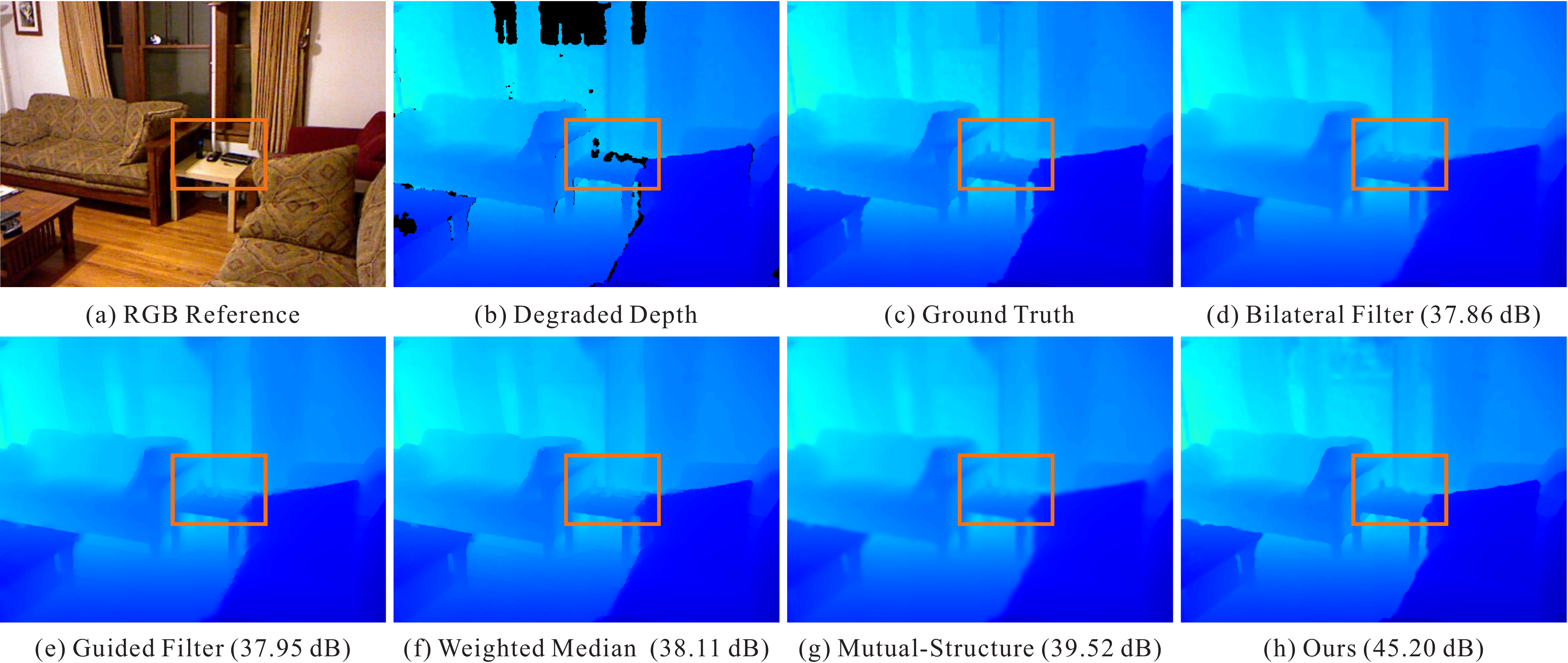}\\
\caption{Comparison of depth/RGB restoration results. The images are from NYU Depth V2
dataset \cite{SilbermanHKF12}. (a) and (b) are the RGB reference image and degraded depth
respectively. (c) is the ground truth. (d-g) are results of bilateral filter
\cite{TomasiM98}, guided image filter \cite{HeST10}, weighted median filter
\cite{ZhangXJ14} and mutual-structure filtering \cite{ShenZXJ15} respectively. (h) is
ours.} \label{fig:rgbdepth}
\end{figure*}

\vspace{0.1in}\noindent\textbf{Depth/RGB Image Restoration~~} Depth map restoration with
guidance of RGB images is an important task since depth captured by cameras such as
Microsoft Kinect, is often with holes and severe noise. Depth estimated by stereo
matching also contains errors in textureless and occlusion regions. Refining the produced
depth is therefore needed for 3D reconstruction and geometry estimation. Our framework
can address this problem.

We evaluate our method on NYU Depth V2 dataset \cite{SilbermanHKF12}, where depth is
captured by Microsoft Kinect. It includes 1,449 aligned depth/RGB pairs. In our
experiment, we treat the provided processed depth maps as ground truth and randomly
select 10\% image pairs for testing and the remaining 90\% for training. We simply stack
the gray-scaled RGB image, depth map, and the mask that indicates location of missing
values as the input to the network. The output is the refined depth map.

Comparison of different methods is reported in Table \ref{tab:depthrgb}. For bilateral
filter \cite{TomasiM98}, guided image filtering \cite{HeST10}, and weighted median filter
\cite{ZhangXJ14}, we first fill in the holes using joint-bilateral-filter-based
normalized convolution \cite{KnutssonW93} and then apply the filters to refine depth. We
implement the method of \cite{LuRL14} given no public code. The lower PSNRs by
filter-based methods in Table \ref{tab:depthrgb} are due to big holes. Methods of Lu
\etal \cite{LuRL14} and mutual-structure filtering \cite{ShenZXJ15} achieve better
results; but they face similar difficulties to restore large regions. our method produces
high-quality results by properly handling different-scale holes.

We also evaluate our framework on the synthetic dataset \cite{LuRL14}. Since there is no
training data provided, we collect it from Middleburry stereo 2014 dataset
\cite{ScharsteinS02} and MPI Sintel stereo dataset \cite{ButlerWSB12}. We degrade depth
maps by adding holes and noise according to the method of \cite{LuRL14}. Our model
setting and training are the same as those on NYU Depth V2 dataset. The results are
listed in Table \ref{tab:depthrgb} where all methods achieve good performance. It is
because the synthetic data is simple. On the one hand, the reference image is with high
quality. On the other hand, holes are relatively small.

In the visual comparison in Figure \ref{fig:rgbdepth}, our result is with sharp
boundaries and many details. More results are provided in our supplementary file.

\begin{table}[t]
\centering
\small
\begin{tabular}{|l|c|c|}
  \hline
  Methods & NYU Depth V2 Data& Lu \etal Data\\
  \hline
  \hline
  Bilateral Filter \cite{TomasiM98}&26.19& 34.33\\
  \hline
  Guided Filter \cite{HeST10} &27.02& 34.86\\
  \hline
  Weighted Median\cite{ZhangXJ14} &31.19&38.17 \\
  \hline
  Lu \etal \cite{LuRL14} &34.53&39.20\\
  \hline
  Mutual-Structure \cite{ShenZXJ15} &33.97& 39.13\\
  \hline
  \textbf{Ours} & \textbf{39.42}& \textbf{39.21}\\
  \hline
\end{tabular}\vspace{0.1in}
\caption{Comparisons of different depth/RGB restoration methods on the NYU Depth V2 and
Lu \etal \cite{LuRL14} datasets. We report average PSNRs of each method on the testing
dataset. }\label{tab:depthrgb}
\end{table}

\begin{figure}[t]
\centering
\begin{tabular}{@{\hspace{0.0mm}}c@{\hspace{1.0mm}}c@{\hspace{1.0mm}}c@{\hspace{0mm}}}
\includegraphics[width=0.32\linewidth]{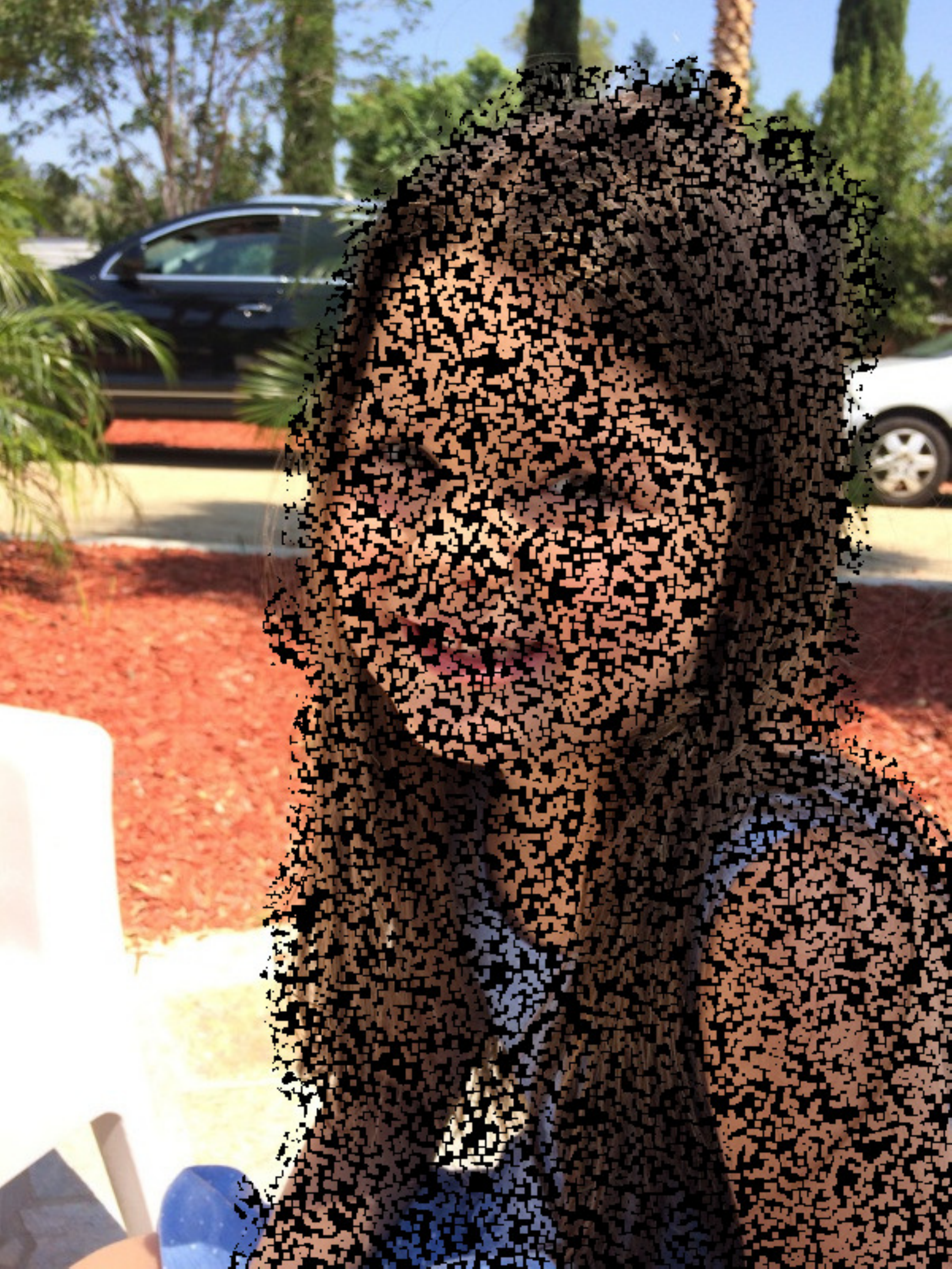}&
\includegraphics[width=0.32\linewidth]{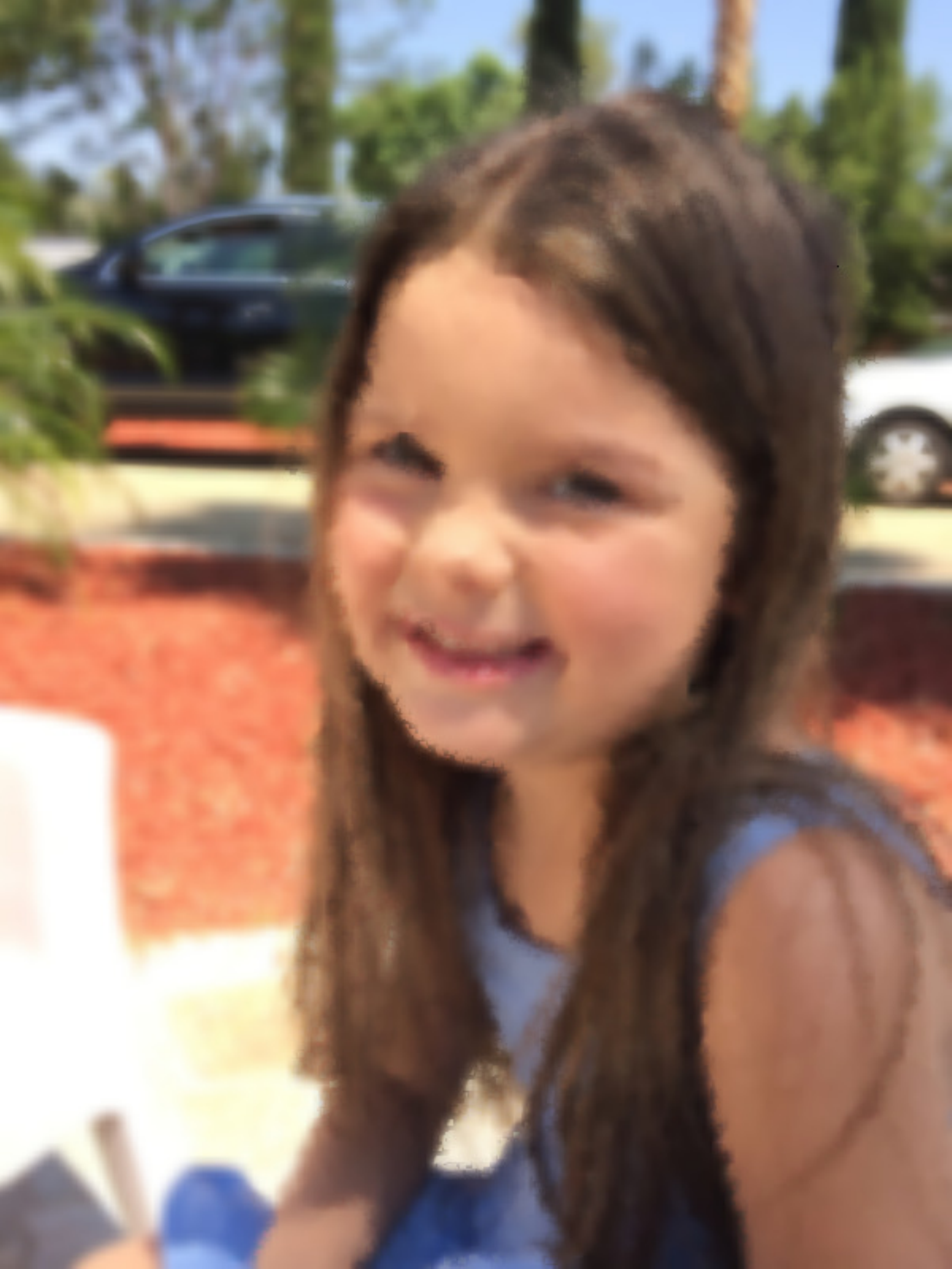}&
\includegraphics[width=0.32\linewidth]{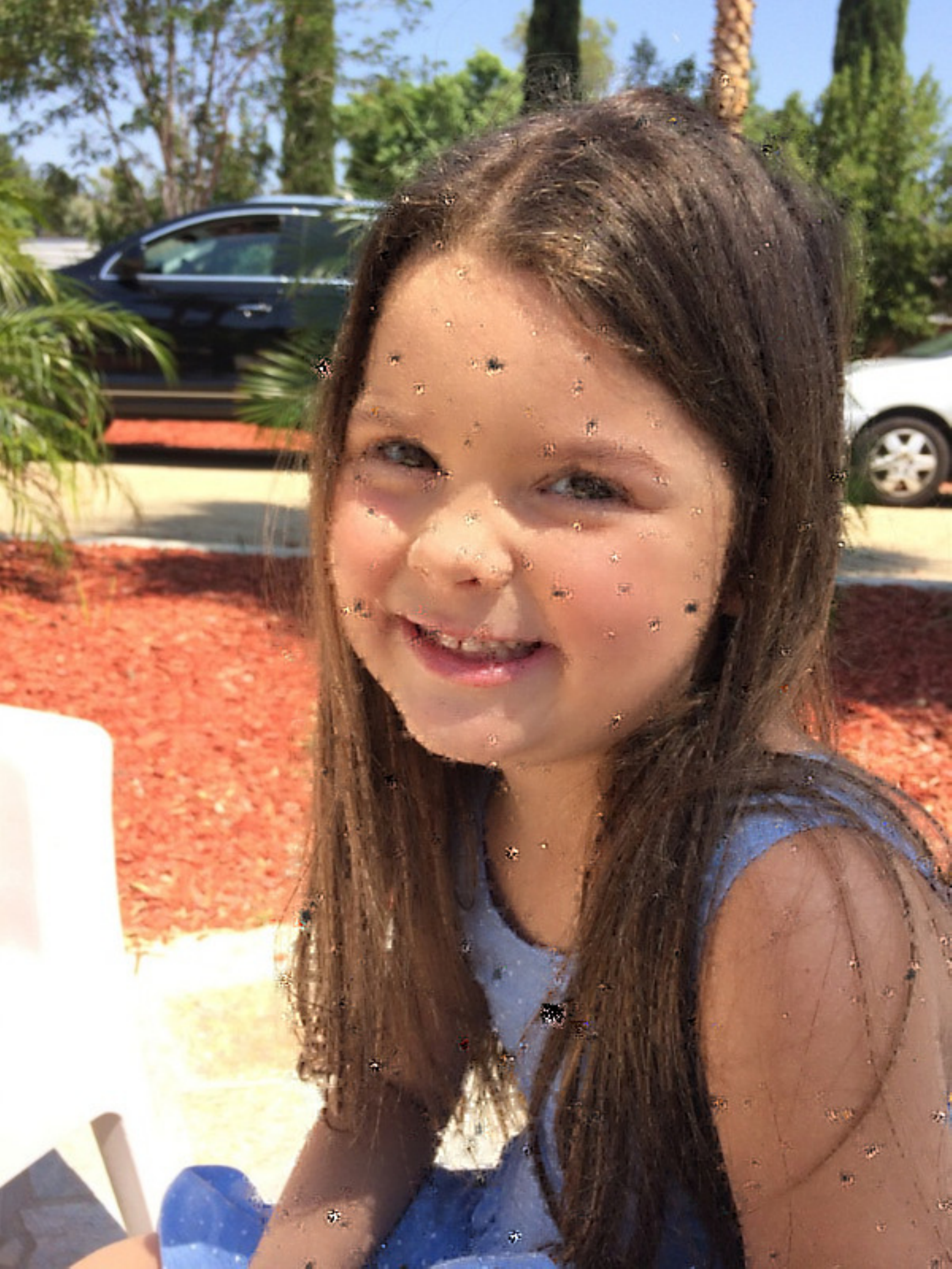}\\
\small (a) Input & \small (b) Result of \cite{KnutssonW93}   & \small (c) Result of \cite{RenXYS15} \\
\includegraphics[width=0.32\linewidth]{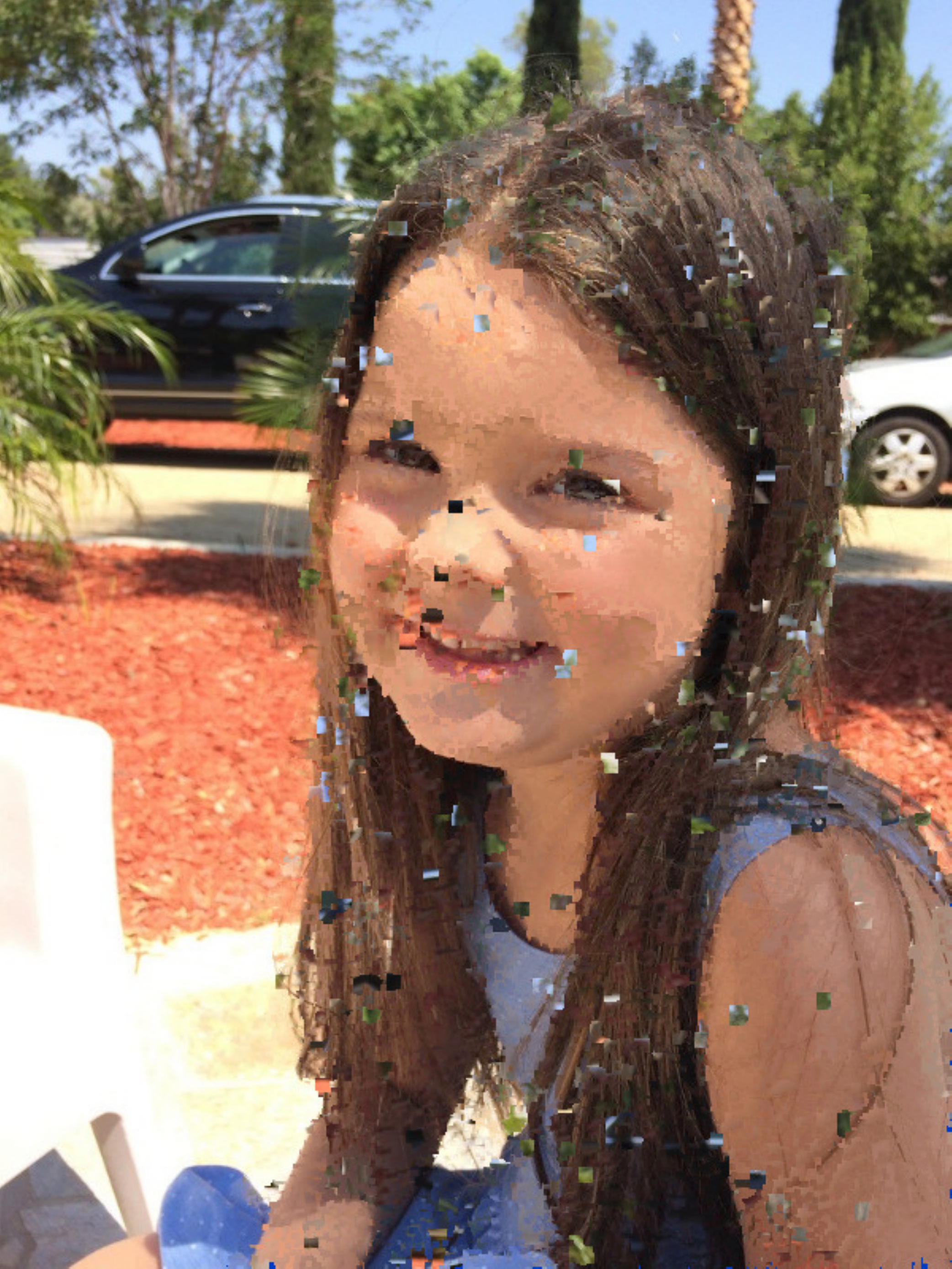}&
\includegraphics[width=0.32\linewidth]{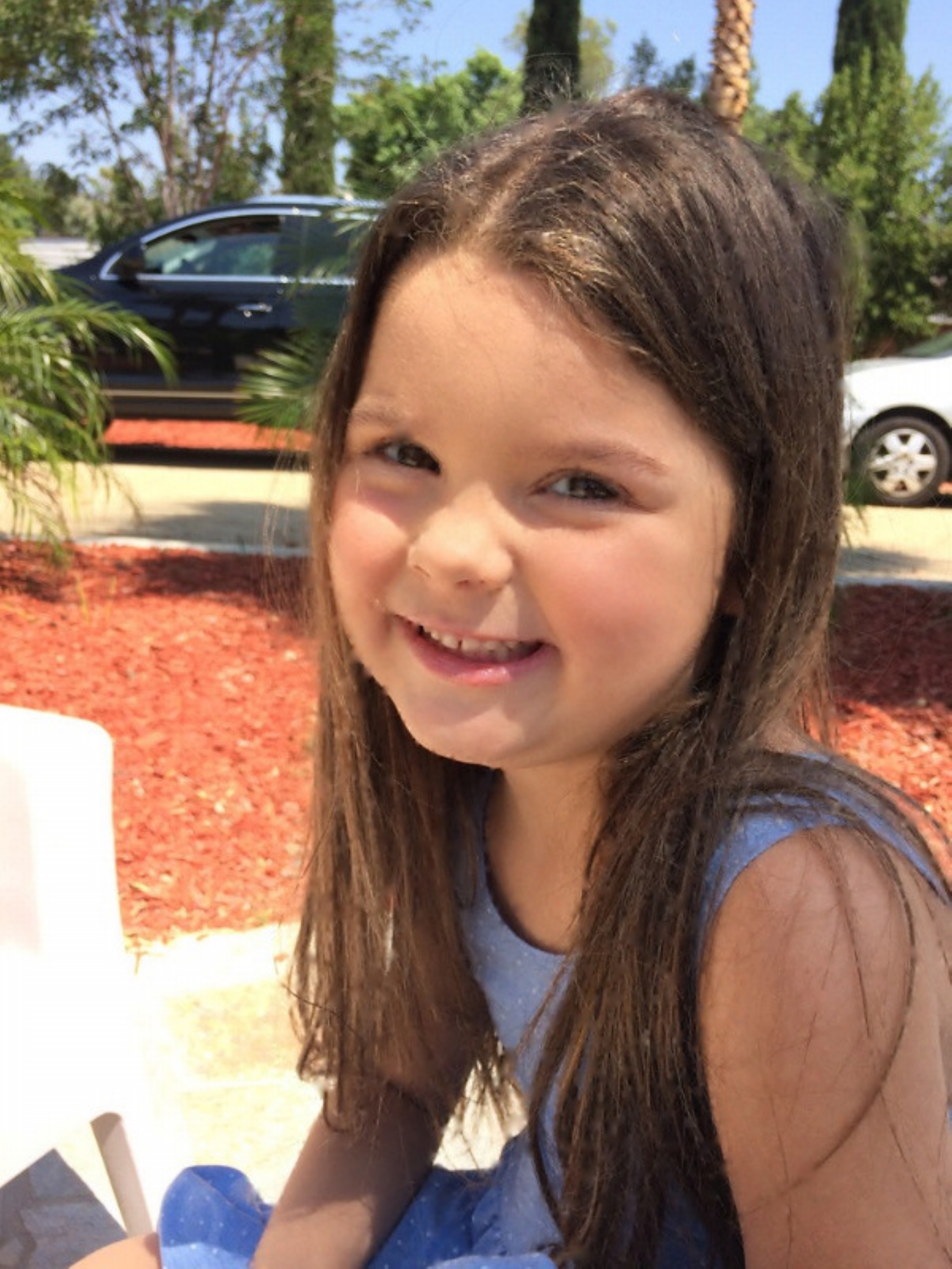}&
\includegraphics[width=0.32\linewidth]{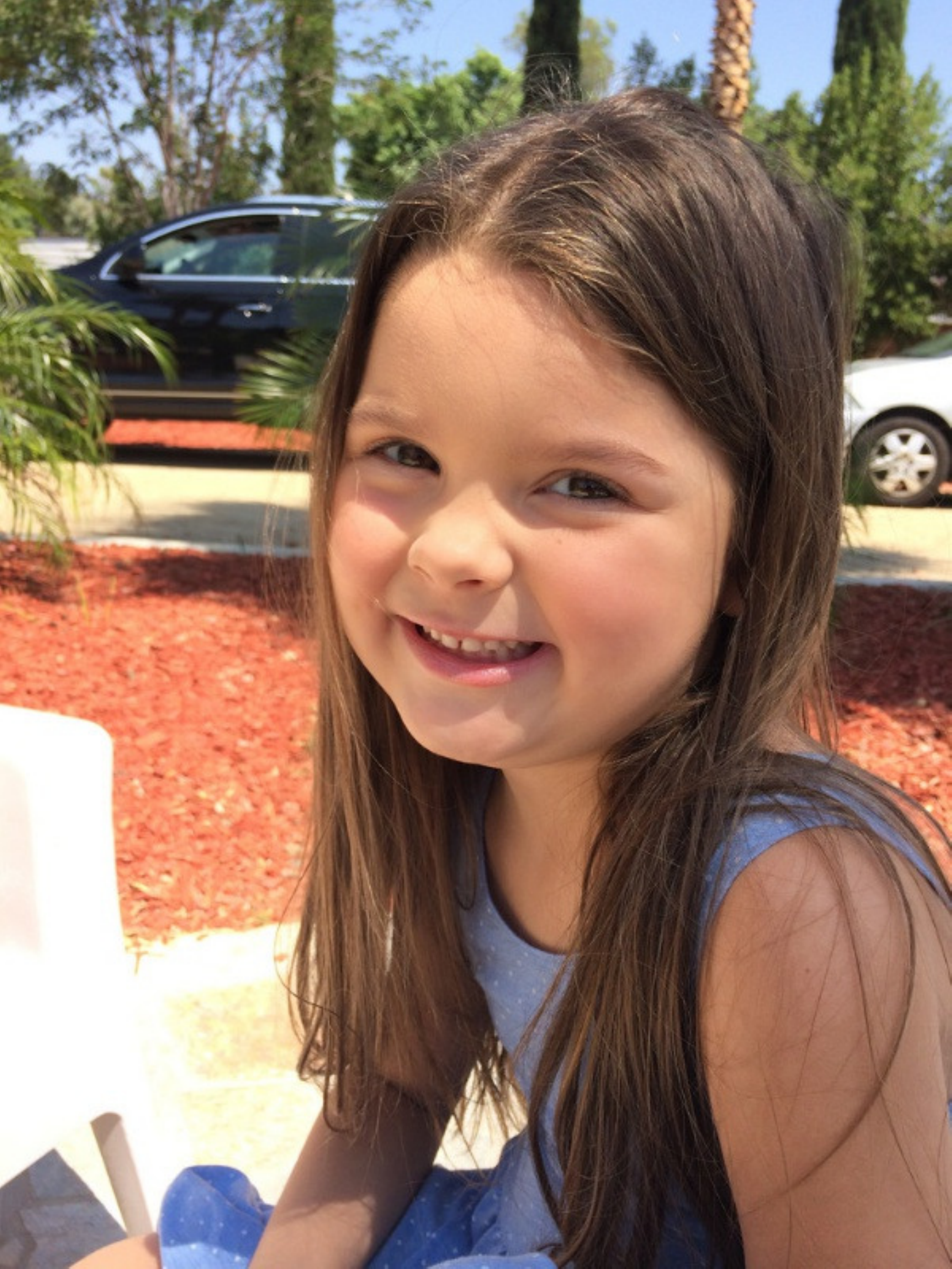}\\
\small (d) Result of \cite{KormanA11} & \small (e) Ours & \small (f) Ground Truth \\
\end{tabular}
\caption{Comparisons of image completion results. (a) is the input. (b-d) are results of
\cite{KnutssonW93,RenXYS15,KormanA11}. (e) is ours and (f) is the ground truth. }
\label{fig:image_completion}
\end{figure}

\vspace{0.1in}\noindent\textbf{Natural Image Completion~~} We apply our method to image
completion. For evaluation, we use the portrait dataset \cite{ShenTGZJ16} since the
portraits are with rich structure and details. For each portrait, we randomly set 5\%
pixels visible and dilate these visible regions by $4$ pixels.
A degraded image is shown as Figure \ref{fig:image_completion}(a).
The input to our framework is the degraded color image. We train our
model with 1,800 portraits from \cite{ShenTGZJ16} with our degradation. One example is
shown in Figure \ref{fig:image_completion} where (a) is the input.
In Figure \ref{fig:image_completion}(b), the result of normalized convolution with
bilateral filter \cite{KnutssonW93} is a bit burry. The result in (c) is our trained
model of \cite{RenXYS15} using our data. This method has difficulty to complete large
holes because of the limited-size receptive field. PatchMatch-based method
\cite{KormanA11} produces result (d). Our result in (e) is a well restored image. The
quantitative evaluation using the testing dataset from \cite{ShenTGZJ16} with our
degradation in Table \ref{tab:img_completion} also manifests our superior performance.

\begin{table}[t]
\centering
\small
\begin{tabular}{|l|c|}
  \hline
  Methods~~& ~~~~PSNRs~~~~ \\
  \hline
  \hline
  Normalized Convolution \cite{KnutssonW93} & 16.05\\
  \hline
  CNNs-based Inpaiting \cite{RenXYS15} & 30.52\\
  \hline
  PatchMatch based Completion \cite{KormanA11}~~~~~~~~~~~~~~~~~~~& 24.81\\
  \hline
  \textbf{Ours} & \textbf{41.21}\\
  \hline
\end{tabular}\vspace{0.1in}
\caption{Evaluation of image completion results on our dataset.
}\label{tab:img_completion}
\end{table}

\vspace{0.1in}\noindent\textbf{Learning Image Filter~~} Our framework also benefits image
filter learning. Similar to those of \cite{XuRYLJ15} and \cite{LiuP016}, we use our deep
convolution pyramid to learn common edge-preserving image filters, including weighted
least square filter (WLS) \cite{FarbmanFLS08}, $L_0$ \cite{XuLXJ11}, RTV \cite{XuYXJ12},
weighted median filter (WMF) \cite{ZhangXJ14}, and rolling guidance filter (RGF)
\cite{ZhangSXJ14}. These filters are representative ones with either global optimization
or local computation. We employ the images from Imagenet \cite{RussakovskyDSKS15} to
train our models. The input to our network is color channels and the output is filtered
color images. We quantitatively evaluate our method and previous approaches
\cite{XuRYLJ15,LiuP016} on the testing dataset of \cite{XuRYLJ15} by measuring PSNRs. As
reported in Table \ref{tab:imgfilter}, our framework outperforms that of \cite{XuRYLJ15}
and achieves comparable performance to state-of-the-art method \cite{LiuP016}. It is
noteworthy that our testing is 31\% faster than that of \cite{LiuP016} on the same
hardware.

\begin{table}[t]
\centering
\small
\begin{tabular}{|l|c|c|c|}
  \hline
  Filter Type~~& PSNRs of \cite{XuRYLJ15} & PSNRs of \cite{LiuP016} & Our PSNRs\\
  \hline
  \hline
  WLS \cite{FarbmanFLS08} &36.2& {39.4}& \textbf{39.6}\\
  \hline
  $L_0$ \cite{XuLXJ11} &\textbf{32.8}&30.9& {32.6}\\
  \hline
  RTV \cite{XuYXJ12} &32.1&37.1& \textbf{38.0}\\
  \hline
  WMF \cite{ZhangXJ14} &31.6&34.0& \textbf{39.3}\\
  \hline
  RGF \cite{ZhangSXJ14} &35.9& {42.2}& \textbf{42.6}\\
  \hline
\end{tabular}\vspace{0.1in}
\caption{Performance of our framework for learning image filters. PSNRs of
\cite{XuRYLJ15} and \cite{LiuP016} are cited from paper \cite{LiuP016}.
}\label{tab:imgfilter}
\end{table}

\begin{figure}[t]
\centering
\includegraphics[width=0.98\linewidth]{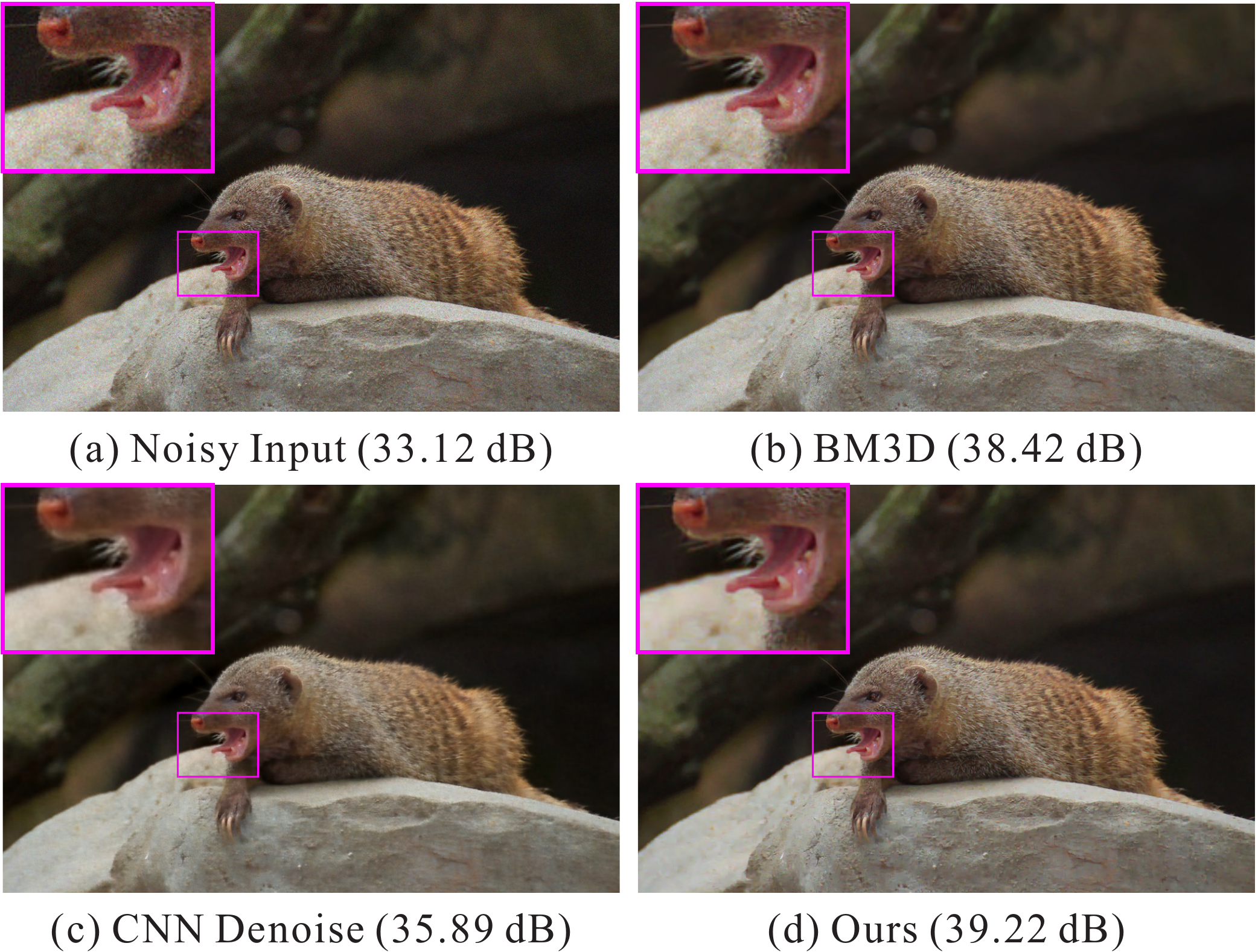}\\
\caption{Comparison of image noise removal. (a) is the noisy input. (b) and (c) are the
results of BM3D \cite{Dabov2007} and CNN based denoising \cite{RenX15} respectively. (d)
is ours. } \label{fig:image_noise_removal}
\end{figure}

\vspace{0.1in}\noindent\textbf{Image Noise Removal~~} Our framework can also be applied
to image noise removal. We train our model on Imagenet data \cite{RussakovskyDSKS15}. To
get the noisy input close to real scenes we add Possion noise and Guassian white noise
with deviation 0.01 to each image. Our model is directly trained on color images. One
example is shown in Figure \ref{fig:image_noise_removal}. Compared with previous BM3D
\cite{Dabov2007} and CNN denoising \cite{RenX15} approaches shown in (b) and (c), our
result contains less noise and well preserved details.

\vspace{0.1in}\noindent\textbf{Other Applications~~} The powerful learning ability of our
framework also benefits other image processing applications of learning image
deconvolution, image colorization, image enhancement, edge refinement \etc More results are provided in
our supplementary file.

\section{Concluding Remarks}

We have proposed a general and efficient convolution neutral network for low-level vision
image processing tasks. It can produce very large receptive fields essential for many
applications while not accordingly increase computation complexity. Our method is based
on pyramid-style learning while introducing new adaptive depth mechanism. We have
provided many examples to verify its effectiveness.

{\small
\bibliographystyle{ieee}
\bibliography{deep_pyramid}
}

\end{document}